\pdfoutput=1

\documentclass[10pt]{article}

\usepackage{Sty/mcr}
\usepackage{Sty/pkg}
\usepackage{Sty/thmE}

\usepackage{url}

\usepackage{subcaption}

\usepackage{Sty/algorithm} 
\usepackage{Sty/algorithmic}

\usepackage{Sty/cancel}

\usepackage{Sty/soul}

\graphicspath{ {./img/} }

\usepackage[margin=1in]{Sty/geometry}
\usepackage{Sty/natbib}

\title{Exact Statistical Inference for the Wasserstein Distance by \\ Selective Inference}

\date{\today}

\author{
    Vo Nguyen Le Duy \\
    Nagoya Institute of Technology and RIKEN\\
    duy.mllab.nit@gmail.com
  \and
    Ichiro Takeuchi\\
    Nagoya Institute of Technology and RIKEN\\
    takeuchi.ichiro@nitech.ac.jp
}

\begin{document}

\maketitle

\begin{abstract}
In this paper, we study statistical inference for the  Wasserstein distance, which has attracted much attention and has been applied to various machine learning tasks.
Several studies have been proposed in the literature,
but almost all of them are based on \emph{asymptotic} approximation and do \emph{not} have finite-sample validity.
In this study, we propose an \emph{exact (non-asymptotic)} inference method for the Wasserstein distance inspired by the concept of conditional Selective Inference (SI).
To our knowledge, this is the first method that can provide a valid confidence interval (CI) for the Wasserstein distance with finite-sample coverage guarantee, which can be applied not only to one-dimensional problems but also to multi-dimensional problems.
We evaluate the performance of the proposed method on both synthetic and real-world datasets.
\end{abstract}

\clearpage


\section{Introduction}

The Wasserstein distance, which is a metric used to compare the probability distributions, has been attracted significant attention and being used more and more in Statistics and Machine Learning (ML) \citep{kolouri2017optimal}.
This distance measures the cost to couple one distribution with another, which arises from the notion of \emph{optimal transport} \citep{villani2009optimal}.
The utilization of the Wasserstein distance benefits to several applications such as supervised learning \citep{frogner2015learning}, generative modelling \citep{arjovsky2017wasserstein}, biology \citep{evans2012phylogenetic}, and computer vision \citep{ni2009local}.

When the Wasserstein distance calculated from noisy data is used for various decision-making problems, it is necessary to quantify its statistical reliability, e.g., in the form of confidence intervals (CIs). 
However, there is no satisfactory statistical inference method for the Wasserstein distance. 
The main reason is that the Wasserstein distance is defined based on the optimal solution of a linear program (LP), and it is difficult to analyze how the uncertainty in the data is transmitted to the uncertainty in the Wasserstein distance.
Several studies have been proposed in literature  \citep{bernton2017inference, del1999tests, del2019central, ramdas2017wasserstein, imaizumi2019hypothesis}.
However, they all rely on asymptotic approximation to mitigate the difficulty stemming from the fact that the Wasserstein distance depends on the optimization problem, and thus most of them can only be applied to one-dimensional problems (the details will be discussed in the related work section).

When an optimization problem such as LP is applied to random data, it is intrinsically difficult to derive the \emph{exact (non-asymptotic)} sampling distribution of the optimal solution. 
In this paper, to resolve the difficulty, we introduce the idea of conditional Selective Inference (SI). 
It is well-known that the optimal solution of an LP depends only on a subset of variables, called \emph{basic variables}. 
Therefore, the LP algorithm can be interpreted as first identifying the basic variables, and then solving the linear system of equations about the basic variables.

Our basic idea is based on the fact that, in the LP problem for the Wasserstein distance, the identification of the basic variables corresponds to the process of determining the \emph{coupling} between the source and destination of the probability mass.
Since the optimal coupling is determined (selected) based on the data, the \emph{selection bias} must be properly considered.
Therefore, to address the selection bias, we propose an exact statistical inference method for the Wasserstein distance by conditioning on the basic variables of the LP, i.e., optimal coupling. 

The main advantage of the proposed method is that it can provide exact (non-asymptotic) CI for the Wasserstein distance, unlike the asymptotic approximations in the existing studies. 
Moreover, while the existing methods are restricted to one-dimensional problems, the proposed method can be directly extended to multi-dimensional problems because the proposed CI is derived based on the fact that the Wasserstein distance depends on the optimal solution of the LP.

%
%

\subsection{Contributions} 
Regarding the high-level conceptual contribution, we introduce a novel approach to explicitly characterize the selection bias of the data-driven coupling problem inspired by the concept of conditional Selective Inference (SI).
In regard to the technical contribution, we provide an exact (non-asymptotic) inference method for the Wasserstein distance.
To our knowledge, this is the first method that can provide \emph{valid} CI, called \emph{selective CI}, for the Wasserstein distance that guarantees the coverage property in finite sample size.
Another practically important advantage of this study is that the proposed method is valid when the Wasserstein distance is computed in multi-dimensional problem, which is impossible for almost all the existing asymptotic methods since the limit distribution of the Wasserstein distance is only applicable for univariate data.
We conduct experiments on both synthetic and real-world datasets to evaluate the performance of the proposed method.

\subsection{Related works} 
In traditional statistics, reliability evaluation with Wasserstein distance has been based on asymptotic theory, i.e., sample size $\rightarrow \infty$.
In the univariate case, instead of solving the optimization problem, the Wasserstein can be described by using an inverse of the distribution function.
For example, let $F^{-1}$ be the quantile function of the data and $F_n^{-1}$  be the empirical quantile function of the generated data, the Wasserstein distance with $\ell_2$ distance is computed by $\int_{0}^{1} (F^{-1}(t) - F_n^{-1}(t))^2 \,dt$.
Based on the quantile function, several studies \citep{del1999tests, del2019central, ramdas2017wasserstein}  derived the asymptotic distribution of the Wasserstein distance.
Obviously, these methods can not guarantee the validity in finite sample size.
Moreover, since the quantile function is only available in univariate case, these methods can not be extended to multivariate cases which are practically important.

Recently, \cite{imaizumi2019hypothesis} has proposed an approach on multidimensional problems. 
However, it is important to clarify that this study does \emph{not} provide statistical inference for the ``original'' Wasserstein distance.
Instead, the authors consider an \emph{approximation} of the Wasserstein distance, which does not require solving a LP.
Besides, this method also relies on asymptotic distribution of the test statistic which is approximated by the Gaussian multiplier bootstrap.
Therefore, to our knowledge, statistical inference method for the Wasserstein distance in multi-dimensional problems is still a challenging open problem.

Conditional SI is a statistical inference framework for correcting selection bias. 
Traditionally, there are mainly two types of approaches for selection bias correction. 
The first approach is family-wise error rate (FWER) control, which includes standard multiple comparison methods such as traditional Bonferroni correction. 
However, the FWER control is too conservative and it is difficult to utilize it for selection bias correction in complex adaptive data analysis. 
Another approach is false discovery rate (FDR) control, in which the target is to control the expected proportion of discoveries that are false at a given significance level. 
The FDR control is less conservative than FWER control, and it is used in many high-dimensional statistical inference problems.

Conditional SI is the third approach for selection bias correction. 
The basic idea of conditional SI was known before, but it becomes popular by the recent seminal work proposed by \cite{lee2016exact}. 
In that study, exact statistical inference on the selected features by Lasso was considered. 
Their basic idea is to employ the sampling distributions of the selected parameter coefficients \emph{conditional on} the selection event that a subset of features is selected by the Lasso. 
By using the conditional distribution in statistical inference, the selection bias of the Lasso (i.e., the fact that the Lasso selected the features based on the data) can be corrected. 
Their contribution was to show that, even in complex data analysis methods such as Lasso, the exact sampling distribution can be characterized if appropriate selection events are considered.

After the seminal work \citep{lee2016exact}, many conditional SI approaches for various feature selection methods were proposed in the literature \citep{loftus2015selective, yang2016selective, tibshirani2016exact, suzumura2017selective}. 
Furthermore, theoretical analyses and new computational methods for conditional SI are still being actively studied \citep{fithian2014optimal, le2021parametric, duy2021more, sugiyama2021more}. 
However, most of conditional SI studies are focused on feature selection problems. 
Although there have been applications to several problems such as change point detection \citep{hyun2018post, duy2020computing, sugiyama2021valid}, outlier detection \citep{chen2019valid, tsukurimichi2021conditional}, and image segmentation \citep{tanizaki2020computing, duy2020quantifying}, these problems can also be interpreted as feature selection in a broad sense. 
Our novelty in this study is to first introduce conditional SI framework for statistical inference on the Wasserstein distance, which is a data-dependent adaptive distance measure. 
Our basic idea is based on the facts that the Wasserstein distance is formulated as the solution of a linear program (LP), and the optimal solution of an LP is characterized by the \emph{selected} basic variables. 
In this study, we consider the sampling distribution of the Wasserstein distance conditional on the selected basic variables, which can be interpreted as considering the selection event on the optimal coupling between the two distributions.

\section{Problem Statement} \label{sec:problem_statement}

To formulate the problem, we consider two vectors corrupted with Gaussian noise as
\begin{align}
	\bm X &= (x_1, ..., x_n)^\top = \bm \mu_{\bm X}  + \bm \veps_{\bm X}, \quad \bm \veps_{\bm X} \sim \NN(\bm 0, \Sigma_{\bm X}), \label{eq:random_X}\\ 
	\bm Y &= (y_1, ..., y_m)^\top = \bm \mu_{\bm Y}  + \bm \veps_{\bm Y}, \quad \bm \veps_{\bm Y} \sim \NN(\bm 0, \Sigma_{\bm Y}) \label{eq:random_Y},
\end{align}
where $n$ and $m$ are the number of instances in each vector,
$\bm \mu_{\bm X}$ and $\bm \mu_{\bm Y}$ are unknown mean vectors, 
$\bm \veps_{\bm X}$ and $\bm \veps_{\bm Y}$ are Gaussian noise vectors with covariances matrices $\Sigma_{\bm X}$ and $\Sigma_{\bm Y}$ assumed to be known or estimable from independent data.
We denote by $P_n$ and $Q_m$ the corresponding empirical measures on $\bm X$ and $\bm Y$.


\subsection{Cost matrix} We define the cost matrix $C(\bm X, \bm Y)$ of pairwise distances ($\ell_1$ distance) between elements of $\bm X$ and $\bm Y$ as 
\begin{align} \label{eq:cost_matrix}
	C(\bm X, \bm Y) 
	& = \big[|x_i - y_j| \big]_{ij} \in \RR^{n \times m}.
\end{align}
We can vectorize $C(\bm X, \bm Y)$ in the form of 
\begin{align} \label{eq:cost_matrix_vec}
\begin{aligned}
	\bm c(\bm X, \bm Y) 
	&= {\rm {vec}} (C(\bm X, \bm Y)) \in \RR^{nm}\\ 
	&= \Theta (\bm X ~ \bm Y)^\top,
\end{aligned}
\end{align}
where  ${\rm {vec}}(\cdot)$ is an operator that transforms a matrix into a vector with concatenated rows.
The matrix $\Theta$ is defined as
\begin{align} \label{eq:matrix_Theta}
	\Theta &=  
	 \cS(\bm X, \bm Y) \circ \Omega  ~ \in \RR^{nm \times (n + m)},  \\ 
	\cS(\bm X, \bm Y) &= {\rm {sign}} \left( \Omega \left (\bm X ~ \bm Y \right)^\top \right)  \in \RR^{nm} , \nonumber \\ 
	\Omega &= 
	\begin{pmatrix}
		\bm 1_m & \bm 0_m  & \cdots & \bm 0_m & - I_m \\ 
		\bm 0_m & \bm 1_m  & \cdots & \bm 0_m & - I_m \\ 
		\vdots & \vdots  & \ddots & \vdots & \vdots \\ 
		\bm 0_m & \bm 0_m  & \cdots & \bm 1_m & - I_m
	\end{pmatrix}
	\in \RR^{nm \times (n + m)}, \nonumber
\end{align}
where the operator $\circ$ is element-wise product, 
${\rm {sign}}(\cdot)$ is the operator that returns an element-wise indication of the sign of a number,
$\bm 1_m \in \RR^m$ is the vector of ones,
$\bm 0_m \in \RR^m$ is the vector of zeros, and $I_m \in \RR^{m \times m}$ is the identity matrix.


\subsection{The Wasserstein distance} To compare two empirical measures 
$P_n$ and $Q_m$ 
with uniform weight vectors 
$\bm 1_n / n$ and $\bm 1_m / m$, 
we consider the following Wasserstein distance, which is defined as the solution of a linear program (LP),
\begin{align} \label{eq:wasserstein}
	W(P_n, Q_m) = \min \limits_{T \in \RR^{n \times m}} & ~ \langle T, C(\bm X, \bm Y)\rangle \\ 
	\text{s.t.} ~~& ~ T \bm{1}_m = \bm 1_n/n, \nonumber \\ 
	& ~ T^\top \bm{1}_n = \bm 1_m/m \nonumber, \\
	& ~ T \geq 0. \nonumber 
\end{align}
Given $\bm X^{\rm {obs}}$ and $\bm Y^{\rm {obs}}$ respectively sampled from models \eq{eq:random_X} and \eq{eq:random_Y} 
\footnote{To make a distinction between random variables and observed variables, we use superscript $^{\rm {obs}}$, e.g., $\bm X$ is a random vector and $\bm X^{\rm {obs}}$ is the observed data vector.},
the Wasserstein distance in (\ref{eq:wasserstein}) on the observed data can be re-written as
\begin{align}  \label{eq:wasserstein_reformulated}
	W(P_n, Q_m) = \min \limits_{\bm t \in \RR^{nm}} ~~ & 
	\bm t^\top \bm c(\bm X^{\rm {obs}}, \bm Y^{\rm {obs}}) \\
	\text{s.t.} ~~
	& S \bm t = \bm h, ~\bm t \geq \bm 0, \nonumber
\end{align}
where $\bm t = {\rm {vec}}(T) \in \RR^{nm}$, 
$\bm c(\bm X^{\rm {obs}}, \bm Y^{\rm {obs}}) \in \RR^{nm}$ is defined in \eq{eq:cost_matrix_vec}, 
$ S = \left( M_r ~ M_c \right )^\top \in \RR^{(n + m) \times nm}$ in which 
\begin{align*}
	M_r = 
	\begin{bmatrix}
		1 ~ \ldots ~  1 & 0 ~ \ldots ~  0 & \ldots & 0 ~ \ldots ~  0 \\
		0 ~ \ldots ~  0 & 1 ~ \ldots ~  1 & \ldots & 0 ~ \ldots ~  0 \\
		 ~ \ldots ~   &  ~ \ldots ~   & \ldots &  ~ \ldots ~   \\
		0 ~ \ldots ~  0 & 0 ~ \ldots ~  0 & \ldots & 1 ~ \ldots ~  1 \\
	\end{bmatrix} \in \RR^{n \times n m}
\end{align*} 
that performs the sum over the rows of $T$ and 
\begin{align*}
	 M_c = 
	\begin{bmatrix}
		I_m & I_m & \ldots & I_m
	\end{bmatrix} \in \RR^{m \times nm}
\end{align*}
that performs the sum over the columns of $T$, and $\bm h = \left (\bm 1_n/n ~ \bm 1_m/m \right)^\top \in \RR^{n + m}$
\footnote{
We note that there always exists exactly one redundant equality constraint in linear equality constraint system in  \eq{eq:wasserstein_reformulated}.
This is due to the fact that sum of all the masses on $\bm X^{\rm {obs}}$ is always equal to sum of all the masses on $\bm Y^{\rm {obs}}$ (i.e., they are all equal to 1).
Therefore, any equality constraint can be expressed as a linear combination of the others, and hence any one constraint can be dropped.
In this paper, we always drop the last equality constraint (i.e., the last row of matrix $S$ and the last element of vector $\bm h$) before solving \eq{eq:wasserstein_reformulated}.
}.


\subsection{Optimal solution and closed-form expression of the distance} 
\begin{figure*}[!t]
\centering
\includegraphics[width=.8\linewidth]{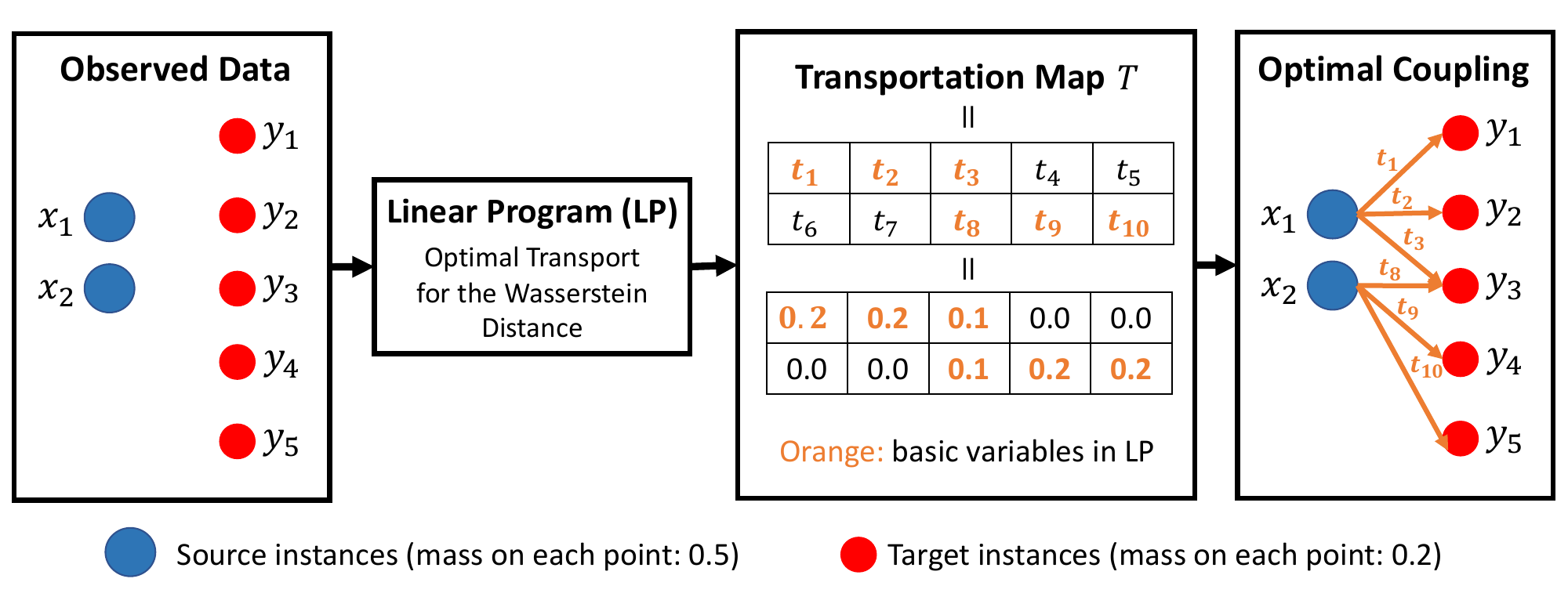}  
\caption{
Illustration of the correspondence between \emph{basic variables} in the LP and the optimal coupling.
After inputting the data to the LP, we obtain the transportation matrix.
The elements $t_1, t_2, t_3, t_8, t_9, t_{10}$ (whose values are non-zero) in the matrix are called basic variables in the LP, and the identification of the basic variables corresponds to the process of determining the coupling between the source and target instances in the optimal transport problem for the Wasserstein distance.
}
\label{fig:illustratrion}
\end{figure*}

Let us denote the set of basis variables (the definition of basis variable can be found in the literature of LP, e.g., \cite{murty1983linear}) obtained when applying the LP in \eq{eq:wasserstein_reformulated} on 
$\bm X^{\rm {obs}}$ and $\bm Y^{\rm {obs}}$ as 
\begin{align} \label{eq:active_set}
	\cM_{\rm {obs}} = \cM(\bm X^{\rm {obs}}, \bm Y^{\rm {obs}}).
\end{align}
%
%
We would like to note that the identification of the basic variables can be interpreted as the process of determining the optimal coupling between the elements of $\bm X^{\rm obs}$ and $\bm Y^{\rm obs}$ in the optimal transport problem for calculating the Wasserstein distance.
Therefore, $\cM_{\rm {obs}}$ in \eq{eq:active_set} can be interpreted as the observed optimal coupling obtained after solving LP in \eq{eq:wasserstein_reformulated} on the observed data \footnote{We suppose that the LP is non-degenerate. A careful discussion might be needed in the presence of degeneracy.}.
An illustration of this interpretation is shown in Figure \ref{fig:illustratrion}.
We also denote by $\cM_{\rm {obs}}^c$ a set of \emph{non-basis variables}.
Then, the optimal solution of \eq{eq:wasserstein_reformulated} can be written as 
\begin{align*}
	\hat{\bm t} \in \RR^{nm}, \quad 
	\hat{\bm t}_{\cM_{\rm {obs}}} = S_{:, \cM_{\rm {obs}}}^{-1}  \bm h, \quad 
	\hat{\bm t}_{\cM_{\rm {obs}}^c} = \bm 0_{|\cM_{\rm {obs}}^c|},
\end{align*} 
where $S_{:, \cM_{\rm {obs}}}$ is a sub-matrix of $S$ made up of all rows and columns in the set $\cM_{\rm {obs}}$.

\begin{example}
Given a matrix
\begin{align*}
	S = 
	\begin{bmatrix}
		1 & 1 & 0 & 0 \\
		0 & 0 & 1 & 1 \\
		1 & 0 & 1 & 0 
	\end{bmatrix}
	\quad \text{and}
	\quad 
	\cM_{\rm {obs}} = \{1, 2, 4\},
\end{align*}
then $S_{:, \cM_{\rm {obs}}}$ is constructed by extracting the first, second and fourth columns of the matrix $S$, i.e,
\begin{align*}
	S_{:, \cM_{\rm {obs}}} =
	\begin{bmatrix}
		1 & 1 & 0 \\
		0 & 0 & 1 \\
		1 & 0 & 0 
	\end{bmatrix}.
\end{align*}
\end{example}

In the literature of LP, the matrix $S_{:, \cM_{\rm {obs}}}$ is also referred to as a \emph{basis}.
After obtaining $\hat{\bm t}$, the Wasserstein distance can be re-written as 
\begin{align} \label{eq:wasserstein_closed_form}
\begin{aligned}
	W(P_n, Q_m) &= \hat{\bm t}^\top  \bm c(\bm X^{\rm {obs}}, \bm Y^{\rm {obs}}) \\ 
	& = \hat{\bm t}_{\cM_{\rm {obs}}}^\top \bm c_{\cM_{\rm {obs}}}(\bm X^{\rm {obs}}, \bm Y^{\rm {obs}}) \\ 
	& =  \hat{\bm t}_{\cM_{\rm {obs}}}^\top 
	\underbrace{
	\Theta_{\cM_{\rm {obs}}, :} (\bm X^{\rm {obs}} ~ \bm Y^{\rm {obs}})^\top
	}_{
	\bm c_{\cM_{\rm {obs}}}(\bm X^{\rm {obs}}, \bm Y^{\rm {obs}})
	},
\end{aligned}
\end{align}
where $\Theta$ is defined in \eq{eq:matrix_Theta},  and $\Theta_{\cM_{\rm {obs}}, :}$ is a sub-matrix of $\Theta$ made up of rows in the set $\cM_{\rm {obs}}$ and all columns.

\subsection{Statistical inference (confidence interval)}
The goal is to provide a confidence interval (CI) for the Wasserstein distance.
The $W(P_n, Q_m)$ in \eq{eq:wasserstein_closed_form} can be written as 
\begin{align} \label{eq:distance_and_eta}
	W(P_n, Q_m) = \bm \eta^\top \left (\bm X^{\rm {obs}} ~ \bm Y^{\rm {obs}} \right )^\top,	
\end{align}
where
$
	\bm \eta = \Theta_{\cM_{\rm {obs}}, :}^\top \hat{\bm t}_{\cM_{\rm {obs}}}
$
is the test-statistic direction.
It is important to note that $\bm \eta$ is a \emph{random} vector because $\cM_{\rm {obs}}$ is selected based on the data.
For the purpose of explanation, let us suppose, for now, that the test-statistic direction $\bm \eta$ in \eq{eq:distance_and_eta} is fixed before observing the data; that is, non-random.
Let us define
\begin{align} \label{eq:tilde_sigma}
	\tilde{\Sigma} = 
	\begin{pmatrix}
		\Sigma_{\bm X} & 0 \\ 
		0 & \Sigma_{\bm Y}
	\end{pmatrix}. 
\end{align}
Thus, we have 
$
	\bm \eta^\top 
	(\bm X ~ \bm Y)^\top
	\sim 
	\NN \left(
	\bm \eta^\top 
	(\bm \mu_{\bm X} ~ \bm \mu_{\bm Y})^\top, 
	\bm \eta^\top \tilde{\Sigma} \bm \eta
	\right )
$.
Given a significance level $\alpha \in [0, 1]$ (e.g., 0.05) for the inference, the \emph{naive} (classical) CI for 
\begin{align*}
	W^{\ast} = W^{\ast}(P_m, Q_m) =  \bm \eta^\top \left (\bm \mu_{\bm X} ~ \bm \mu_{\bm Y} \right )^\top
\end{align*} 
with $(1 - \alpha)$-coverage is obtained by
\begin{align} \label{eq:ci_naive}
	C_{\rm {naive}} = 
	\left \{ 
	w \in \RR : 
	\frac{\alpha}{2} \leq
	F_{w, \sigma^2} 
	\left (
	\bm \eta^\top 
	{\bm X^{\rm {obs}} \choose \bm Y^{\rm {obs}}}
	 \right )
	\leq 1 - \frac{\alpha}{2} 
	\right \}, 
\end{align}
where 
$
\sigma^2 = 
\bm \eta^\top 
\tilde{\Sigma}
\bm \eta
$
and 
$F_{w, \sigma^2}(\cdot)$
is the c.d.f of the normal distribution with a mean $w$ and variance $\sigma^2$.
With the assumption that $\bm \eta$ in \eq{eq:distance_and_eta} is fixed in advance, the naive CI is valid in the sense that 
\begin{align} \label{eq:ci_naive_validity}
	\PP\left( W^\ast \in C_{\rm {naive}} \right ) = 1 - \alpha.
\end{align}
However, in reality, because the test-statistic direction $\bm \eta$ is actually not fixed in advance, the naive CI in \eq{eq:ci_naive} is \emph{unreliable}.
In other words, the naive CI is \emph{invalid} in the sense that the $(1 - \alpha)$-coverage guarantee in \eq{eq:ci_naive_validity} is no longer satisfied.
This is because the construction of $\bm \eta$ depends on the data and selection bias exists.

In the next section, we introduce an approach to correct the selection bias which is inspired by the conditional SI framework, and propose a valid CI called selective CI ($C_{\rm {sel}}$) for the Wasserstein distance which guarantees the $(1 - \alpha)$-coverage property in finite sample (i.e., non-asymptotic).

\section{Proposed Method} \label{sec:proposed_method}

We present the technical details of the proposed method in this section. 
We first introduce the selective CI for the Wasserstein distance in \S \ref{subsec:selective_ci}.
To compute the proposed selective CI, we need to consider the sampling distribution of the Wasserstein distance conditional on the selection event.
Thereafter, the selection event is explicitly characterized in a conditional data space whose identification is presented in \S \ref{subsec:conditional_data_space}.
Finally, we end the section with the detailed algorithm.

\subsection{Selective Confidence Interval for Wasserstein Distance} \label{subsec:selective_ci}

In this section, we propose an exact (non-asymptotic) selective CI for the Wasserstein distance by conditional SI.
We interpret the computation of the Wasserstein distance in \eq{eq:wasserstein_reformulated} as a two-step procedure:
\begin{itemize}
	\item \textbf{Step 1}: Compute the cost matrix in \eq{eq:cost_matrix} with the $\ell_1$ distance and obtain the vectorized form of the cost matrix in \eq{eq:cost_matrix_vec}.
	\item \textbf{Step 2}: Solving the LP in \eq{eq:wasserstein_reformulated} to obtain $\cM_{\rm {obs}}$ which is subsequently used to construct the test-statistic direction $\bm \eta$ in \eq{eq:distance_and_eta} and calculate the distance.
\end{itemize}
Since the distance is computed in data-driven fashion, 
for constructing a valid CI,
it is necessary to remove the information that has been used for the initial calculation process, 
i.e., steps 1 and 2 in the above procedure, 
to correct the selection bias.
This can be achieved by considering the sampling distribution of the test-statistic $\bm \eta^\top (\bm X ~ \bm Y)^\top$ conditional on the selection event; that is,
%
\begin{align} \label{eq:conditional_inference}
	\bm \eta^\top  {\bm X  \choose \bm Y}
	 \Big|  
	\left \{ 
	 \cS(\bm X, \bm Y) = \cS(\bm X^{\rm {obs}}, \bm Y^{\rm {obs}}),
	  \cM(\bm X, \bm Y) = \cM_{\rm {obs}}
	\right \} , 
\end{align}
where $\cS(\bm X, \bm Y)$ is the sign vector explained in the construction of $\Theta$ in \eq{eq:matrix_Theta}. 

\textbf{Interpretation of the selection event in \eq{eq:conditional_inference}.} 
The first and second conditions in \eq{eq:conditional_inference} respectively represent the selection event for steps 1 and 2 in the procedure described in the beginning of this section. 
The first condition 
$\cS(\bm X, \bm Y) = \cS(\bm X^{\rm {obs}}, \bm Y^{\rm {obs}})$
indicates that ${\rm {sign}}(x_i - y_j)  = {\rm {sign}}(x^{\rm {obs}}_i - y_j^{\rm {obs}})$ for $i \in [n], j \in [m]$.
In other words, for $i \in [n], j \in [m]$, we condition on the event whether $x^{\rm {obs}}_i$ is greater than $y^{\rm {obs}}_j$ or not.
The second condition 
$\cM(\bm X, \bm Y) = \cM_{\rm {obs}}$
indicates that the set of selected basic variables for random vectors $\bm X$ and $\bm Y$ is the same as that for $\bm X^{\rm {obs}}$ and $\bm Y^{\rm {obs}}$.
This condition can be interpreted as conditioning on the observed optimal coupling $\cM_{\rm {obs}}$ between the elements of $\bm X^{\rm {obs}}$ and $\bm Y^{\rm {obs}}$, which is obtained after solving the LP in \eq{eq:wasserstein_reformulated} on the observed data (see Figure \ref{fig:illustratrion}).

\textbf{Selective CI.} 
To derive exact statistical inference for the Wasserstein distance, we introduce so-called selective CI for 
$
W^{\ast} = W^{\ast}(P_m, Q_m) =  \bm \eta^\top \left (\bm \mu_{\bm X} ~ \bm \mu_{\bm Y} \right )^\top
$
that satisfies the following $(1 - \alpha)$-coverage property conditional on the selection event:
\begin{align} \label{eq:ci_selective_validity}
	\PP\left( W^\ast \in C_{\rm {sel}} ~ \Big| ~ 
		\cS(\bm X, \bm Y) = \cS(\bm X^{\rm {obs}}, \bm Y^{\rm {obs}}),		
		\cM(\bm X, \bm Y) = \cM_{\rm {obs}}	
	\right ) = 1 - \alpha,
\end{align}
for any $\alpha \in [0, 1]$. 
The selective CI is defined as 
\begin{align} \label{eq:ci_selective}
	C_{\rm {sel}} = 
	\left \{ 
	w \in \RR : 
	\frac{\alpha}{2} \leq
	F_{w, \sigma^2}^{\cZ} 
	\left (
	\bm \eta^\top 
	{\bm X^{\rm {obs}} \choose \bm Y^{\rm {obs}} }
	\right )
	\leq 1 - \frac{\alpha}{2} 
	\right \}.
\end{align}
where the pivotal quantity
\begin{align} \label{eq:pivotal_quantity}
	F_{w, \sigma^2}^{\cZ}
	\left (
	\bm \eta^\top {\bm X \choose \bm Y}
	\right )
	\bigg |
	\left \{ 
	\cS(\bm X, \bm Y) = \cS(\bm X^{\rm {obs}}, \bm Y^{\rm {obs}}), 
	\cM(\bm X, \bm Y) = \cM_{\rm {obs}},  
	\bm q(\bm X, \bm Y) = \bm q(\bm X^{\rm {obs}}, \bm Y^{\rm {obs}})
	\right \}
\end{align}
is the c.d.f of the \emph{truncated} normal distribution with a mean 
$w \in \RR$, 
variance 
$\sigma^2 = \bm \eta^\top \tilde{\Sigma}\bm \eta$,
and truncation region $\cZ$ (the detailed construction of $\cZ$ will be discussed later in \S \ref{subsec:conditional_data_space}) which is calculated based on the selection event in \eq{eq:pivotal_quantity}.
The $\bm q(\bm X, \bm Y)$ in the additional third condition is the sufficient statistic of nuisance parameter defined as
\begin{align*} 
	\bm q(\bm X, \bm Y) = 
	\Big ( I_{n+m} - \bm c \bm \eta^\top \Big ) 
	\Big ( \bm X ~ \bm Y \Big )^\top 	
\end{align*}
in which 
$\bm c = \tilde{\Sigma} \bm \eta (\bm \eta^\top \tilde{\Sigma} \bm \eta)^{-1}$
with
$\tilde{\Sigma}$ is defined in \eq{eq:tilde_sigma}.
Here, we note that the selective CI depends on $\bm q(\bm X, \bm Y)$ because the pivotal quantity in \eq{eq:pivotal_quantity} depends on this component, but the sampling property in \eq{eq:ci_selective_validity} is kept satisfied without this additional condition because we can marginalize  over all values of $\bm q(\bm X, \bm Y)$.
The $\bm q(\bm X, \bm Y)$ corresponds to the component $\bm z$ in the seminal paper of \cite{lee2016exact} (see Section 5, Eq. 5.2 and Theorem 5.2).
We note that additionally conditioning on $\bm q(\bm X, \bm Y)$ is a standard approach in the SI literature and it is used in almost all the SI-related works that we cited in this paper.

To obtain the selective CI in \eq{eq:ci_selective}, we need to compute the quantity in \eq{eq:pivotal_quantity} which depends on the truncation region $\cZ$.
Therefore, the remaining task is to identify $\cZ$ whose characterization will be introduced in the next section.

\subsection{Conditional Data Space Characterization} \label{subsec:conditional_data_space}

We define the set of $(\bm X ~ \bm Y)^\top \in \RR^{n + m}$ that satisfies the conditions in Equation \eq{eq:pivotal_quantity} as 
\begin{align} \label{eq:conditional_data_space}
	\cD = 
	\left \{ 
		{\bm X \choose \bm Y} \in \RR^{n + m}
		 ~ \bigg |  ~ 
		\cS(\bm X, \bm Y) = \cS(\bm X^{\rm {obs}}, \bm Y^{\rm {obs}}), 
		\cM(\bm X, \bm Y) = \cM_{\rm {obs}}, 
		\bm q(\bm X, \bm Y) = \bm q(\bm X^{\rm {obs}}, \bm Y^{\rm {obs}})
	\right \}. 
\end{align}
According to the third condition 
$\bm q(\bm X, \bm Y) = \bm q(\bm X^{\rm {obs}}, \bm Y^{\rm {obs}})$, the data in $\cD$ is restricted to a line in $\RR^{n + m}$ as stated in the following Lemma.
\begin{lemma} \label{lemma:data_line}
Let us define
\begin{align} \label{eq:a_b_line}
	\bm a = 
	\bm q(\bm X^{\rm {obs}}, \bm Y^{\rm {obs}}) 
	\quad \text{and} \quad 
	\bm b = \tilde{\Sigma} \bm \eta (\bm \eta^\top \tilde{\Sigma} \bm \eta)^{-1}, 
\end{align}
where $\tilde{\Sigma}$ is defined in \eq{eq:tilde_sigma}.
Then, the set $\cD$ in \eq{eq:conditional_data_space} can be rewritten using the scalar parameter $z \in \RR$ as follows:
\begin{align} \label{eq:conditional_data_space_line}
	\cD = \Big \{ (\bm X ~ \bm Y)^\top = \bm a + \bm b z \mid z \in \cZ \Big \},
\end{align}
where 
\begin{align} \label{eq:cZ}
	\cZ = \left \{ 
	z \in \RR ~
	\Big |  
	\begin{array}{l}
	\cS(\bm a + \bm b z) = \cS(\bm X^{\rm {obs}}, \bm Y^{\rm {obs}}), \\ 
	\cM(\bm a + \bm b z) = \cM_{\rm {obs}}
	\end{array}
	\right \}.
\end{align}
Here, with a slight abuse of notation, 
$
\cS(\bm a + \bm b z) = \cS \left ((\bm X ~ \bm Y)^\top \right)
$
is equivalent to $\cS(\bm X, \bm Y)$.
This similarly applies to $\cM(\bm a + \bm b z)$.
\end{lemma}

\begin{proof}
According to the third condition in \eq{eq:conditional_data_space}, we have 
\begin{align*}
	\bm q(\bm X, \bm Y) = &~\bm q(\bm X^{\rm {obs}}, \bm Y^{\rm {obs}}) \\ 
	\Leftrightarrow 
	\Big ( I_{n+m} - \bm c \bm \eta^\top \Big ) 
	{ \bm X \choose \bm Y }
	= 
	&~\bm q(\bm X^{\rm {obs}}, \bm Y^{\rm {obs}}) \\ 
	\Leftrightarrow 
	{ \bm X \choose \bm Y }
	= 
	\bm q(\bm X^{\rm {obs}}, \bm Y^{\rm {obs}})
	&+ \frac{\tilde{\Sigma} \bm \eta}{\bm \eta^\top \tilde{\Sigma} \bm \eta}
	\bm \eta^\top  
	{ \bm X \choose \bm Y }.
\end{align*}
By defining 
$\bm a = \bm q(\bm X^{\rm {obs}}, \bm Y^{\rm {obs}})$,
$\bm b = \tilde{\Sigma} \bm \eta (\bm \eta^\top \tilde{\Sigma} \bm \eta)^{-1}$,
$z = \bm \eta^\top  \Big ( \bm X ~ \bm Y \Big )^\top$, and incorporating the first and second conditions in \eq{eq:conditional_data_space}, we obtain the results in Lemma \ref{lemma:data_line}. 
We note that the fact of restricting the data to the line has been already implicitly exploited in the seminal conditional SI work of \cite{lee2016exact}, but explicitly discussed for the first time in Section 6 of \cite{liu2018more}.
\end{proof}

Lemma \ref{lemma:data_line} indicates that we do not have to consider the $(n + m)$-dimensional data space.
Instead, we only to consider the \emph{one-dimensional projected} data space $\cZ$ in \eq{eq:cZ}, which is the truncation region that is important for computing the pivotal quantity in \eq{eq:pivotal_quantity} and constructing the selective CI $C_{\rm {sel}}$ in \eq{eq:ci_selective}.


\textbf{Characterization of truncation region $\cZ$.}
We can decompose $\cZ$ into two separate sets as 
$\cZ = \cZ_1 \cap \cZ_2$,  where 
\begin{align*}
	\cZ_1 &= \{ z \in \RR \mid  \cS(\bm a + \bm b z) = \cS(\bm X^{\rm {obs}}, \bm Y^{\rm {obs}})\} \\ 
	\quad \text{ and } \quad 
	\cZ_2 &= \{ z \in \RR \mid \cM(\bm a + \bm b z) = \cM_{\rm {obs}} \}.
\end{align*}
We first present the construction of $\cZ_1$ in the following Lemma.

\begin{lemma} \label{lemma:cZ_1}
For notational simplicity, we denote $\bm s_{\rm {obs}} = \cS(\bm X^{\rm {obs}}, \bm Y^{\rm {obs}})$. 
Then, the $\cZ_1$ is an interval defined as:
\begin{align*}
	\cZ_1 = \left \{ z \in \RR \mid 
	\max \limits_{j: \nu_j^{(2)} > 0} \frac{ - \nu_j^{(1)}}{\nu_j^{(2)}}
	\leq z \leq
	\min \limits_{j: \nu_j^{(2)} < 0} \frac{ - \nu_j^{(1)}}{\nu_j^{(2)}}
	\right \},
\end{align*}
where
$\bm \nu^{(1)} = \bm s_{\rm {obs}}  \circ  \Omega \bm a$ 
and 
$\bm \nu^{(2)} = \bm s_{\rm {obs}}  \circ  \Omega \bm b$.
\end{lemma}
\begin{proof}
Let us first remind that 
$\cS(\bm X, \bm Y) = {\rm {sign}} \left( \Omega \left (\bm X ~ \bm Y \right)^\top \right)$ where $\Omega$ is defined in \eq{eq:cost_matrix_vec}.
Then, the $\cZ_1$ can be re-written as follows:
\begin{align*}
	\cZ_1 &= \{ z \in \RR \mid  \cS(\bm a + \bm b z) = \bm s_{\rm {obs}}\}  \\ 
	 &= \left \{ z \in \RR \mid  {\rm {sign}}\Big( \Omega (\bm a + \bm b z) \Big) = \bm s_{\rm {obs}} \right \}  \\ 
	&= \left \{ z \in \RR \mid  \bm s_{\rm {obs}}  \circ  \Omega (\bm a + \bm b z) \geq \bm 0\right \}.
\end{align*}
By defining $\bm \nu^{(1)} = \bm s_{\rm {obs}}  \circ  \Omega \bm a$
and  
$\bm \nu^{(2)} = \bm s_{\rm {obs}}  \circ  \Omega \bm b$, the result of Lemma \ref{lemma:cZ_1} is straightforward
by solving the above system of linear inequalities.
\end{proof}

Next, we present the construction of $\cZ_2$. Here,  $\cZ_2$ can be interpreted as the set of values of $z$ in which we obtain the same set of the selected basic variables $\cM_{\rm {obs}}$ when applying the LP in \eq{eq:wasserstein_reformulated} on the prametrized data $\bm a + \bm b z$.

\begin{lemma} \label{lemma:cZ_2}
The set $\cZ_2$ is an interval defined as:
\begin{align*}
	\cZ_2 = 
	\left \{ 
	z \in \RR \mid 
	\max \limits_{j \in \cM_{\rm {obs}}^c: \tilde{v}_j > 0} \frac{ - \tilde{u}_j}{\tilde{v}_j}
	\leq z \leq
	\min \limits_{j \in \cM_{\rm {obs}}^c: \tilde{v}_j < 0} \frac{ - \tilde{u}_j}{\tilde{v}_j}
	\right \},
\end{align*}
where 
\begin{align*}
\tilde{\bm u} &= 
\left(
	\bm u_{\cM^c}^\top - \bm u_{\cM}^\top S_{:, \cM}^{-1} S_{:, \cM^c}
\right)^\top,
\\ 
\tilde{\bm v} &= 
\left (
	\bm v_{\cM^c}^\top - \bm v_{\cM}^\top S_{:, \cM}^{-1} S_{:, \cM^c}
\right )^\top , 	
\end{align*}
$\bm u = \Theta \bm a$, $\bm v = \Theta \bm b$, $\Theta$ is defined as in \eq{eq:matrix_Theta} with observed data.
\end{lemma}

\begin{proof}
We consider the LP in \eq{eq:wasserstein_reformulated} with the parametrized data $\bm a + \bm b z$ as follows:
\begin{align}  \label{eq:prametric_LP}
	\min \limits_{\bm t \in \RR^{nm}} ~~  
	\bm t^\top 
	\Theta (\bm a + \bm b z) 
	\quad 
	\text{s.t.}
	\quad 
	 S \bm t = \bm h, \bm t \geq \bm 0. 
\end{align}
Here, we remind that $\Theta (\bm a + \bm b z)$ is the vectorized version of the cost matrix defined in \eq{eq:cost_matrix_vec}.
The optimization problem in \eq{eq:prametric_LP} is well-known as the \emph{parametric cost problem} in LP literature (e.g., see Section 8.2 in \cite{murty1983linear}).
Let us denote $\bm u = \Theta \bm a$ and $\bm v = \Theta \bm b$, 
the LP in \eq{eq:prametric_LP} can be re-written as 
\begin{align}  \label{eq:prametric_LP_simplified}
	\min \limits_{\bm t \in \RR^{nm}} ~~ 
	(\bm u + \bm v z)^\top \bm t 
	\quad 
	\text{s.t.}
	\quad 
	S \bm t = \bm h, \bm t \geq \bm 0.
\end{align}
Given a fixed value $z_0$, let $\cM$ be an optimal basic index set of the LP in \eq{eq:prametric_LP_simplified} at $z = z_0$ and $\cM^c$ be its complement.
Then by partitioning $S$, $\bm t$, $\bm u$, and $\bm v$ as 
\begin{align*}
	S &= [S_{:, \cM}, S_{:, \cM^c}] \\ 
	\bm t = (\bm t_{\cM}, \bm t_{\cM^c}), \quad 
	\bm u &= (\bm u_{\cM}, \bm u_{\cM^c}), \quad 
	\bm v = (\bm v_{\cM}, \bm v_{\cM^c}),
\end{align*}
the LP in \eq{eq:prametric_LP_simplified} becomes
\begin{align} \label{eq:LP_decompose}
	\min \limits_{\bm t_{\cM}, \bm t_{\cM^c}} ~~ &
	(\bm u_{\cM} + \bm v_{\cM} z)^\top \bm t_{\cM}
	+  
	(\bm u_{\cM^c} + \bm v_{\cM^c} z)^\top \bm t_{\cM^c} \nonumber
	\\
	\text{s.t.} ~~
	& S_{:, \cM} \bm t_{\cM} + 
	S_{:, \cM^c} \bm t_{\cM^c} = \bm h,  \\ 
	& \bm t_{\cM} \geq \bm 0, ~ \bm t_{\cM^c} \geq \bm 0. \nonumber
\end{align}
The value of $\bm t_{\cM}$ can be computed as 
\begin{align*}
	\bm t_{\cM} = 
	S_{:, \cM}^{-1} \bm h 
	- S_{:, \cM}^{-1}S_{:, \cM^c} \bm t_{\cM^c},
\end{align*}
and this general expression when substituted in the objective (cost) function of \eq{eq:prametric_LP_simplified} yields
%
%
\begin{align*}
	f &= 
	(\bm u_{\cM} + \bm v_{\cM} z)^\top 
	(S_{:, \cM}^{-1} \bm h - S_{:, \cM}^{-1}S_{:, \cM^c} \bm t_{\cM^c})  +  
	(\bm u_{\cM^c} + \bm v_{\cM^c} z)^\top \bm t_{\cM^c}\\ 
	& = 
	(\bm u_{\cM} + \bm v_{\cM} z)^\top 
	S_{:, \cM}^{-1} \bm h  +
	\Big [
	\left(
	\bm u_{\cM^c}^\top - \bm u_{\cM}^\top S_{:, \cM}^{-1} S_{:, \cM^c}
	\right) + 
	 \left (
	\bm v_{\cM^c}^\top - \bm v_{\cM}^\top S_{:, \cM}^{-1} S_{:, \cM^c}
	\right ) \times z
	\Big ] 
	\bm t_{\cM^c},
\end{align*}
which expresses the cost of \eq{eq:LP_decompose} in terms of $\bm t_{\cM^c}$.
Let us denote 
\begin{align*}
\tilde{\bm u} = 
\left(
	\bm u_{\cM^c}^\top - \bm u_{\cM}^\top S_{:, \cM}^{-1} S_{:, \cM^c}
\right)^\top
\text{ and } ~~
\tilde{\bm v} = 
\left (
	\bm v_{\cM^c}^\top - \bm v_{\cM}^\top S_{:, \cM}^{-1} S_{:, \cM^c}
\right )^\top , 	
\end{align*}
we can write $\bm r_{\cM^c} = \tilde{\bm u} + \tilde{\bm v} z$ which is known as the \emph{relative cost vector} in the LP literature.
%
%
%
Then, $\cM$ is an optimal basic index set of \eq{eq:prametric_LP_simplified} for all values of the parameter $z$ satisfying 
\begin{align} \label{eq:criterion_optimal_basis}
	\bm r_{\cM^c} = \tilde{\bm u} + \tilde{\bm v} z \geq 0,
\end{align}
which is also explicitly discussed in Section 8.2 of \cite{murty1983linear}.
Finally, the results in Lemma \ref{lemma:cZ_2} are obtained by respectively replacing $\cM$ and $\cM^c$ by $\cM_{\rm {obs}}$ and $\cM_{\rm {obs}}^c$, and solving the linear inequality system in \eq{eq:criterion_optimal_basis}.
\end{proof}

Once $\cZ_1$ and $\cZ_2$ are identified, we can compute the truncation region $\cZ = \cZ_1 \cap \cZ_2$.
Finally, we can use $\cZ$ to calculate the pivotal quantity in \eq{eq:pivotal_quantity} which is subsequently used to 
construct the proposed selective CI in \eq{eq:ci_selective}.
The details of the algorithm is presented in Algorithm \ref{alg:selective_ci_wasserstein_distance}.

\begin{algorithm}[!t]
\renewcommand{\algorithmicrequire}{\textbf{Input:}}
\renewcommand{\algorithmicensure}{\textbf{Output:}}
 \begin{algorithmic}[1]
  \REQUIRE $\bm X^{\rm {obs}}, \bm Y^{\rm {obs}}$
  \vspace{3pt}
  \STATE Compute the cost matrix as in \eq{eq:cost_matrix} and obtained its vectorized version 
  $\bm c(\bm X^{\rm {obs}}, \bm Y^{\rm {obs}})$ as in \eq{eq:cost_matrix_vec}
  \vspace{3pt}
  \STATE Solve LP in \eq{eq:wasserstein_reformulated} to obtain $\cM_{\rm {obs}}$
  \vspace{3pt}
  \STATE Compute $\bm \eta$ based on $\cM_{\rm {obs}}$ $\leftarrow$ Equation \eq{eq:distance_and_eta}
  \vspace{3pt}
  \STATE Calculate $\bm a$ and $\bm b$ based on $\bm \eta$ $\leftarrow$ Equation \eq{eq:a_b_line} 
  \vspace{3pt}
  \STATE Construct $\cZ_1$ $\leftarrow$ Lemma \ref{lemma:cZ_1}
  \vspace{3pt}
  \STATE Identify $\cZ_2$ $\leftarrow$ Lemma \ref{lemma:cZ_2}
  \vspace{3pt}
  \STATE Truncation region $\cZ = \cZ_1 \cap \cZ_2$
  \vspace{3pt}
  \STATE $C_{\rm {sel}}$ $\leftarrow$ Equation \eq{eq:ci_selective}
  \vspace{3pt}
  \ENSURE $C_{\rm {sel}}$
 \end{algorithmic}
\caption{Selective CI for the Wasserstein Distance}
\label{alg:selective_ci_wasserstein_distance}
\end{algorithm}


\section{Extension to Multi-Dimension} \label{sec:extension}

In \S \ref{sec:problem_statement} and  \S \ref{sec:proposed_method}, we mainly focus on the Wasserstein distance in one-dimension, i.e., $x_{i \in [n]} \in \RR$ and $y_{j \in [m]} \in \RR$.
In this section, we generalize the problem setup and extend the proposed method for the Wasserstein distance in multi-dimension.
We consider two random sets $X$ and $Y$ of $d$-dimensional vectors 
\begin{align} \label{eq:random_data_gen}
	\begin{aligned}
	  X &= (\bm x_{1, :}, \ldots, \bm x_{n, :})^\top  \in \RR^{n \times d},  
	\\ 
	Y &= (\bm y_{1, :}, \ldots, \bm y_{m, :})^\top \in  \RR^{m \times d} ,
	\end{aligned}
\end{align}
corrupted with Gaussian noise as 
\begin{align*}
	  \RR^{nd} \ni \bm X_{\rm {vec}} &= {\rm {vec}} (X) = (\bm x_{1, :}^\top,  \ldots, \bm x_{n, :}^\top ) ^\top 
	  = \bm \mu_{\bm X_{\rm {vec}}}  + \bm \veps_{\bm X_{{\rm {vec}}}}, 
	  \quad \bm \veps_{\bm X_{\rm {vec}}} \sim \NN(\bm 0, \Sigma_{\bm X_{{\rm {vec}}}}),  \\ 
	   \RR^{md} \ni \bm Y_{\rm {vec}} &= {\rm {vec}}(Y) = (\bm y_{1, :}^\top,  \ldots, \bm y_{n, :}^\top ) ^\top 
	  = \bm \mu_{\bm Y_{\rm {vec}}}  + \bm \veps_{\bm Y_{{\rm {vec}}}}, 
	  \quad \bm \veps_{\bm Y_{\rm {vec}}} \sim \NN(\bm 0, \Sigma_{\bm Y_{{\rm {vec}}}}), 
\end{align*}
where $n$ and $m$ are the number of instances in each set,
$\bm \mu_{\bm X_{\rm {vec}}}$ and $\bm \mu_{\bm Y_{\rm {vec}}}$ are unknown mean vectors, 
$\bm \veps_{\bm X_{\rm {vec}}}$ and $\bm \veps_{\bm Y_{\rm {vec}}}$ are Gaussian noise vectors with covariances matrices $\Sigma_{\bm X_{\rm {vec}}}$ and $\Sigma_{\bm Y_{\rm {vec}}}$ assumed to be known or estimable from independent data.


\textbf{Cost matrix.}
The cost matrix $C(X, Y)$ of pairwise distances ($\ell_1$ distance) between elements of $X$ and $Y$ as 
\begin{align} \label{eq:cost_matrix_gen}
	C(X, Y) 
	& = \left[ \sum \limits_{k=1}^d |\bm x_{i, k} - \bm y_{j, k}| \right]_{ij} \in \RR^{n \times m}.
\end{align}
Then, the vectorized form of $C(X, Y) $ can be defined as 
\begin{align} \label{eq:cost_matrix_vec_gen}
\begin{aligned}
	 \bm c(X, Y)
	&= 
	{\rm {vec}} \left( C(X, Y) \right) \\ 
	&= 
	\Theta^{\rm {mul}} \left (\bm X_{\rm {vec}} ~ \bm Y_{\rm {vec}} \right)^\top  \in \RR^{nm},
\end{aligned}
\end{align}
where 
\begin{align*}
	\Theta^{\rm {mul}} &=  
	\sum \limits_{k=1}^d \cS_k(X, Y) \circ
	\left( \Omega \otimes  \bm e_{d, k}^\top \right) \in \RR^{nm \times (nd + md)},  \\ 
	\cS_k(X, Y) &= {\rm {sign}} \left( \left( \Omega \otimes  \bm e_{d, k}^\top \right) \left (\bm X_{\rm {vec}} ~ \bm Y_{\rm {vec}} \right)^\top \right) \in \RR^{nm},
\end{align*}
the matrix $\Omega$ is defined in \eq{eq:matrix_Theta},  
the operator $\otimes$ is Kronecker product, 
and $\bm e_{d, k} \in \RR^d$ is a $d$-dimensional unit vector with 1 at position $k \in [d]$.


\textbf{The Wasserstein distance in multi-dimension.} 
Given $X^{\rm {obs}}$ and $Y^{\rm {obs}}$   sampled from \eq{eq:random_data_gen}, 
after obtaining $\bm c(X^{\rm {obs}}, Y^{\rm {obs}})$ as in \eq{eq:cost_matrix_vec_gen},
the Wasserstein distance in multi-dimension is defined as 
\begin{align}  \label{eq:wasserstein_reformulated_gen}
	W^{\rm {mul}} (P_n, Q_m) = \min \limits_{\bm t \in \RR^{nm}} ~~ & 
	\bm t^\top \bm c(X^{\rm {obs}}, Y^{\rm {obs}}) \\
	\text{s.t.} ~~
	& S \bm t = \bm h, ~\bm t \geq \bm 0, \nonumber
\end{align}
where $S$ and $\bm h$ are defined in \eq{eq:wasserstein_reformulated}.
By solving LP in \eq{eq:wasserstein_reformulated_gen}, we obtain the set of selected basic variables 
\begin{align} \label{eq:active_set_gen}
	\cM_{\rm {obs}} = \cM(X^{\rm {obs}}, Y^{\rm {obs}}),
\end{align}
Then, the Wasserstein distance can be re-written as 
\begin{align}
\begin{aligned}
	W^{\rm {mul}}(P_n, Q_m) &= 
	\hat{\bm t}^\top  \bm c(X^{\rm {obs}}, Y^{\rm {obs}}) \\ 
	& = \hat{\bm t}_{\cM_{\rm {obs}}}^\top \bm c_{\cM_{\rm {obs}}}( X^{\rm {obs}},  Y^{\rm {obs}}) \\ 
	& =  \hat{\bm t}_{\cM_{\rm {obs}}}^\top 
	\Theta^{\rm {mul}}_{\cM_{\rm {obs}}, :} (\bm X^{\rm {obs}}_{\rm {vec}} ~ \bm Y^{\rm {obs}}_{\rm {vec}})^\top \\
	&= \bm \eta^\top_{\rm {mul}} (\bm X^{\rm {obs}}_{\rm {vec}} ~ \bm Y^{\rm {obs}}_{\rm {vec}})^\top
\end{aligned}
\end{align}
where 
$
\bm \eta_{\rm {mul}} = 
	\left (\hat{\bm t}_{\cM_{\rm {obs}}}^\top 
	\Theta^{\rm {mul}}_{\cM_{\rm {obs}}, :} \right )^\top 
$
is the test-statistic direction,
$
\hat{\bm t}_{\cM_{\rm {obs}}} = 
	S_{:, \cM_{\rm {obs}}}^{-1}  \bm h
$ is the optimal solution of \eq{eq:wasserstein_reformulated_gen},
and the matrix 
$\Theta^{\rm {mul}}$ is defined in \eq{eq:cost_matrix_vec_gen}.


\textbf{Selection event and selective CI.} 
Since we are dealing with multi-dimensional case, the selection event is slightly different from but more general than the one presented in \eq{eq:pivotal_quantity} of \S \ref{sec:proposed_method}.
Specifically, we consider the following conditional inference 
\begin{align} \label{eq:conditional_inference_gen}
	\bm \eta^\top_{\rm {mul}}  (\bm X_{\rm {vec}}  ~ \bm Y_{\rm {vec}})^\top 
	  \mid  
	 \cE^{\rm {mul}},
\end{align}
where 
\begin{align*}
	\cE^{\rm {mul}} = \left \{ 
	\bigcup_{k = 1}^d \cS_k(X, Y) = \cS_k(X^{\rm {obs}}, Y^{\rm {obs}}), 
	 \cM(X, Y) = \cM_{\rm {obs}}, 
	\bm q(X, Y) = \bm q(X^{\rm {obs}}, Y^{\rm {obs}})
	\right \}.
\end{align*}
Once the selection event $\cE^{\rm {mul}}$ has been identified, the pivotal quantity can be computed:
\begin{align*} 
	F_{\bm \eta^\top_{\rm {mul}} 
	(\bm \mu_{\bm X_{\rm {vec}}} ~ \bm \mu_{\bm Y_{\rm {vec}}})^\top, \sigma^2_{\rm {mul}}}^{\cZ^{\rm {mul}}}
	\left (
	\bm \eta^\top_{\rm {mul}} (\bm X_{\rm {vec}} ~ \bm Y_{\rm {vec}})^\top
	\right )
	\mid 
	\cE^{\rm {mul}}
\end{align*}
where 
$\sigma^2_{\rm {mul}} = \bm \eta_{\rm {mul}}^\top \tilde{\Sigma}^{\rm {mul}} \bm \eta_{\rm {mul}}$ with
$
\tilde{\Sigma}^{\rm {mul}}  = 
\begin{pmatrix}
	\Sigma_{\bm X_{\rm {vec}}} & 0 \\ 
	0 & \Sigma_{\bm Y_{\rm {vec}}}
\end{pmatrix}
$,
and truncation region $\cZ^{\rm {mul}}$ is calculated based on the selection event $\cE^{\rm {mul}}$ which we will discuss later.
After $\cZ^{\rm {mul}}$ is identified, the selective CI is defined as 
\begin{align} \label{eq:ci_selective_gen}
	C^{\rm {mul}}_{\rm {sel}} = 
	\left \{ 
	w \in \RR: 
	\frac{\alpha}{2} \leq
	F_{w, \sigma^2_{\rm {mul}}}^{\cZ^{\rm {mul}}} 
	\left (\bm \eta^\top_{\rm {mul}} 
	{\bm X^{\rm {obs}}_{\rm {vec}} \choose \bm Y^{\rm {obs}}_{\rm {vec}} }
	\right )
	\leq 1 - \frac{\alpha}{2} 
	\right \}.
\end{align}
The remaining task is to identify $\cZ^{\rm {mul}}$.


\textbf{Characterization of truncation region $\cZ^{\rm {mul}}$.}
Similar to the discussion in \S \ref{sec:proposed_method}, the data is restricted on the line due to the conditioning on the nuisance component $\bm q(X, Y)$.
Then, the set of data that satisfies the condition in \eq{eq:conditional_inference_gen} is defined as 
\begin{align*} 
	\cD^{\rm {mul}} = \Big \{ (\bm X_{\rm {vec}} ~ \bm Y_{\rm {vec}})^\top = \bm a^{\rm {mul}} + \bm b^{\rm {mul}} z \mid z \in \cZ^{\rm {mul}} \Big \},
\end{align*}
where 
\begin{align*}
	\bm a^{\rm {mul}} &= \bm q(X^{\rm {obs}}, Y^{\rm {obs}}),
	\\ 
	\bm b^{\rm {mul}} &= \tilde{\Sigma}^{\rm {mul}} \bm \eta_{\rm {mul}} (\bm \eta^\top_{\rm {mul}} \tilde{\Sigma}^{\rm {mul}} \bm \eta_{\rm {mul}})^{-1},
	\\ 
	\cZ^{\rm {mul}} &= \left \{ 
	z  \in \RR
	~ \Bigg | ~ 
	\begin{array}{l}
	\bigcup \limits_{k = 1}^d
	\cS_k(\bm a^{\rm {mul}} + \bm b^{\rm {mul}} z) = \cS_k(X^{\rm {obs}}, Y^{\rm {obs}}), 
	\\ 
	\cM(\bm a^{\rm {mul}} + \bm b^{\rm {mul}} z) = \cM_{\rm {obs}}
	\end{array}
	\right \}
\end{align*}
with $z \in \RR$.
Next, we can decompose $\cZ^{\rm {mul}}$ into two separate sets as 
$\cZ^{\rm {mul}} = \cZ_1^{\rm {mul}} \cap \cZ_2^{\rm {mul}}$,  where 
\begin{align*}
	\cZ_1^{\rm {mul}} & = \left \{ z \in \RR \mid  \bigcup \limits_{k = 1}^d \cS_k(\bm a^{\rm {mul}} + \bm b^{\rm {mul}} z) = \cS_k(X^{\rm {obs}}, Y^{\rm {obs}})\right \}, \\  
	\cZ_2^{\rm {mul}} & = \left \{ z \in \RR \mid \cM(\bm a^{\rm {mul}} + \bm b^{\rm {mul}} z) = \cM_{\rm {obs}} \right \}.
\end{align*}
From now, the identification of $\cZ_1^{\rm {mul}}$ and $\cZ_2^{\rm {mul}}$ is straightforward and similar to the construction of $\cZ_1$ and $\cZ_2$ discussed in \S \ref{sec:proposed_method}.
Once $\cZ_1^{\rm {mul}}$ and $\cZ_2^{\rm {mul}}$ are identified, we can compute the truncation region $\cZ^{\rm {mul}} = \cZ_1^{\rm {mul}} \cap \cZ_2^{\rm {mul}}$ and use it to compute the selective CI in \eq{eq:ci_selective_gen}.

\section{Experiment} \label{sec:experiment}

In this section, we demonstrate the performance of the proposed method in both univariate case and multi-dimensional case. 
We present the results on synthetic data in \S \ref{subsec:numerical}.
Thereafter, the results on real data are shown in \S \ref{subsec:real_data}.
In all the experiments, we set the significance level $\alpha = 0.05$, i.e., all the experiments were conducted with the coverage level of $1 - \alpha = 0.95$.

\subsection{Numerical Experiment} \label{subsec:numerical}

In this section, we evaluate the performance of the proposed selective CI in terms of coverage guarantee, CI length and computational cost.
We also show the results of comparison between our selective CI and the naive CI in \eq{eq:ci_naive} in terms of coverage guarantee.
We would like to note that we did not conduct the comparison in terms of CI length because the naive CI could not guarantee the coverage property.
In statistical viewpoint, if the CI is unreliable, i.e., invalid or does not satisfy the coverage property, the demonstration of CI length does not make sense.
Besides, we additionally compare the performance of the proposed method with the latest asymptotic study \citep{imaizumi2019hypothesis}.


\begin{figure}[!t]

\begin{subfigure}{.33\linewidth}
  \centering
  \includegraphics[width=\linewidth]{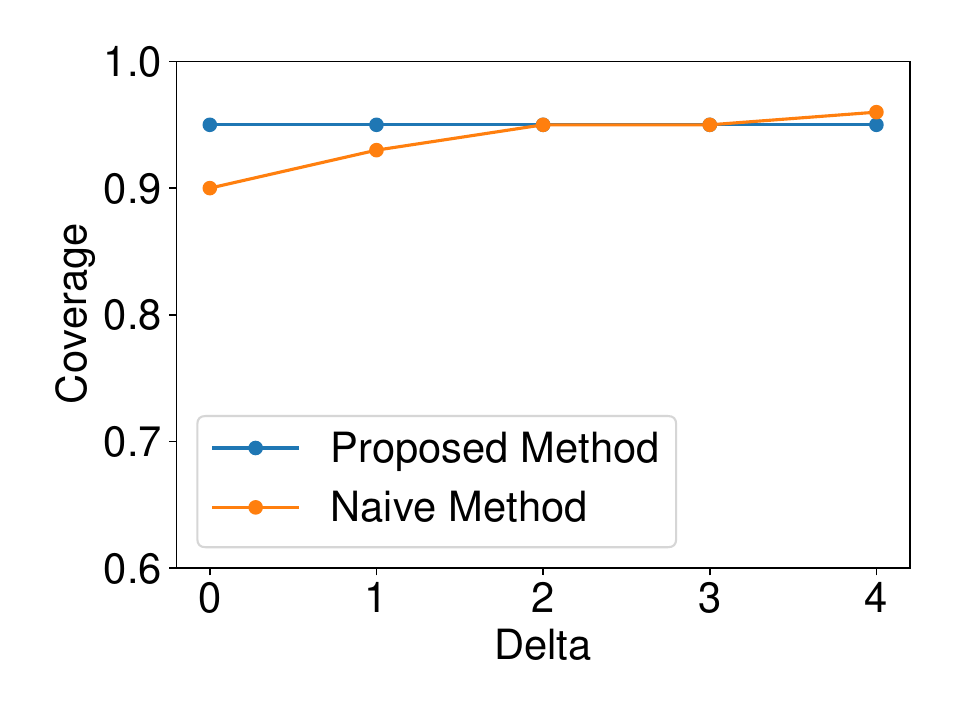}  
  \caption{Coverage guarantee}
\end{subfigure}
\begin{subfigure}{.33\linewidth}
  \centering
  \includegraphics[width=\linewidth]{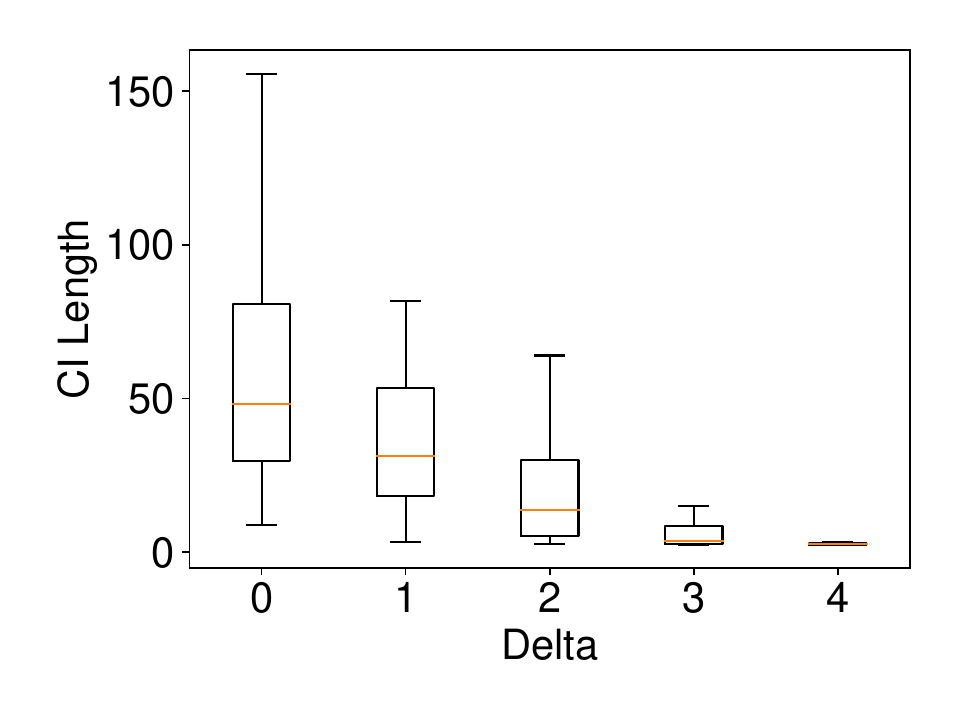}  
  \caption{CI length}
\end{subfigure}
\begin{subfigure}{.33\linewidth}
  \centering
  \includegraphics[width=\linewidth]{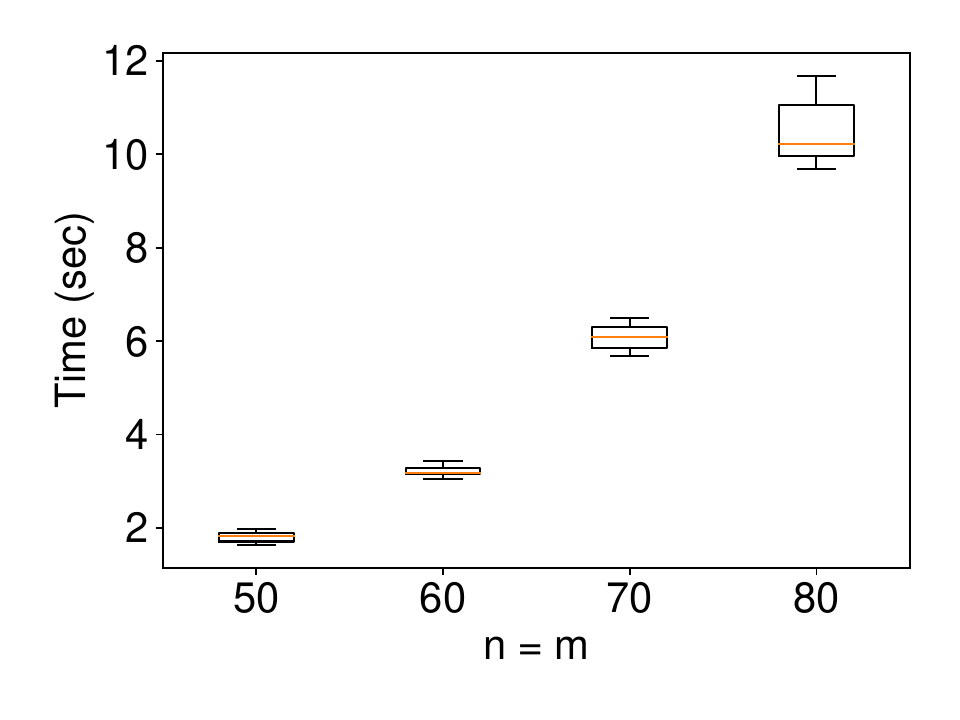}  
  \caption{Computational cost}
\end{subfigure}
\caption{Coverage guarantee, CI length and computational cost in univariate case ($d = 1$).}
\label{fig:numerical_1_d}
\end{figure}

\subsubsection{Univariate case ($d = 1$)}
We generated the dataset $\bm X$ and $\bm Y$ with 
$\bm \mu_{\bm X} = \bm 1_n$, 
$\bm \mu_{\bm Y} = \bm 1_m + \Delta$ (element-wise addition),
$\bm \veps_{\bm X} \sim \NN(\bm 0, I_n)$,
$\bm \veps_{\bm Y} \sim \NN(\bm 0, I_m)$.
Regarding the experiments of coverage guarantee and CI length, we set $n = m = 5$ and ran 120 trials for each $\Delta \in \{0, 1, 2, 3, 4\}$.
In regard to the experiments of computational cost, we set $\Delta = 2$ and ran 10 trials for each $n = m \in \{50, 60, 70, 80\}$. 
The results are shown in Figure \ref{fig:numerical_1_d}.
In the left plot, the naive CI can not properly guarantee the coverage property while the proposed selective CI does.
The results in the middle plot indicate that the larger the true distance between $\bm X$ and $\bm Y$, the shorter selective CI we obtain.
The right plot shows that the proposed method also has reasonable computational cost.


\begin{figure}[!t]

\begin{subfigure}{.33\linewidth}
  \centering
  \includegraphics[width=\linewidth]{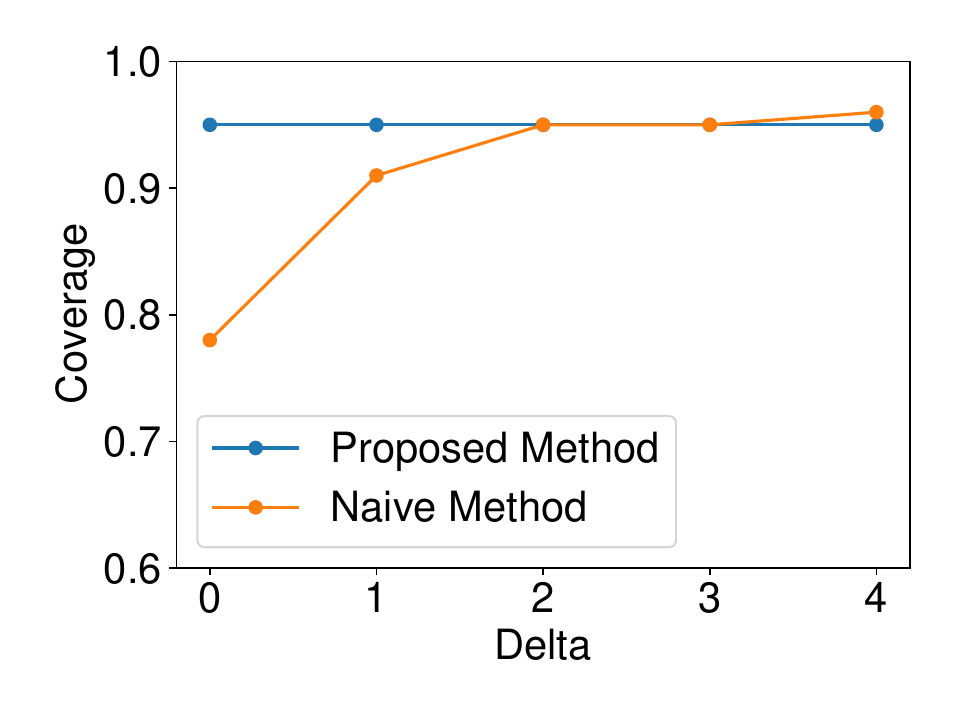}  
  \caption{Coverage guarantee}
\end{subfigure}
\begin{subfigure}{.33\linewidth}
  \centering
  \includegraphics[width=\linewidth]{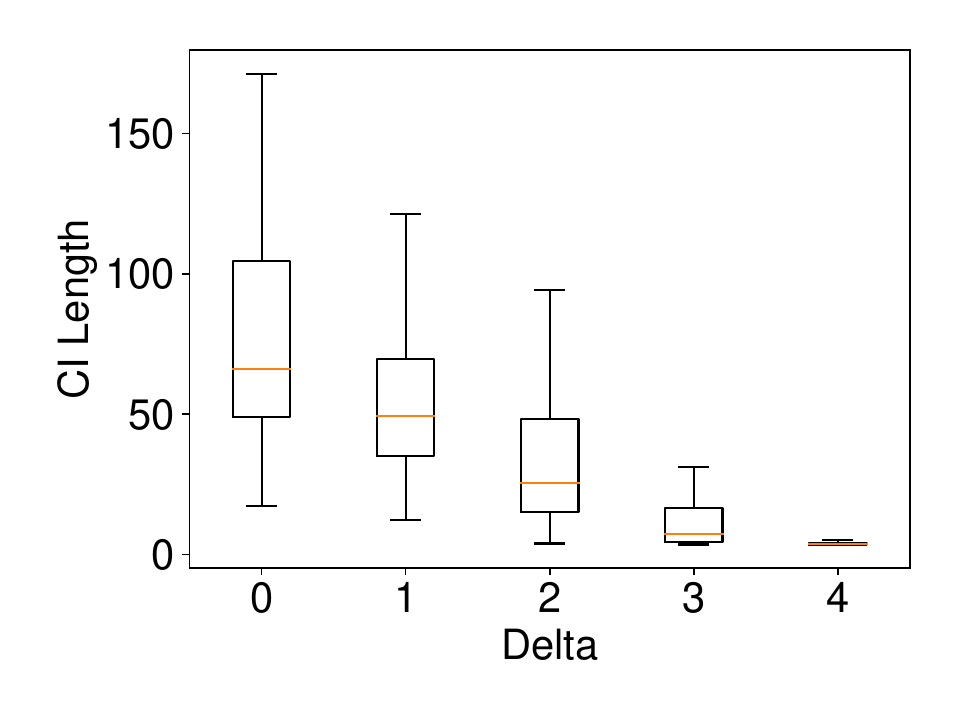}  
  \caption{CI length}
\end{subfigure}
\begin{subfigure}{.33\linewidth}
  \centering
  \includegraphics[width=\linewidth]{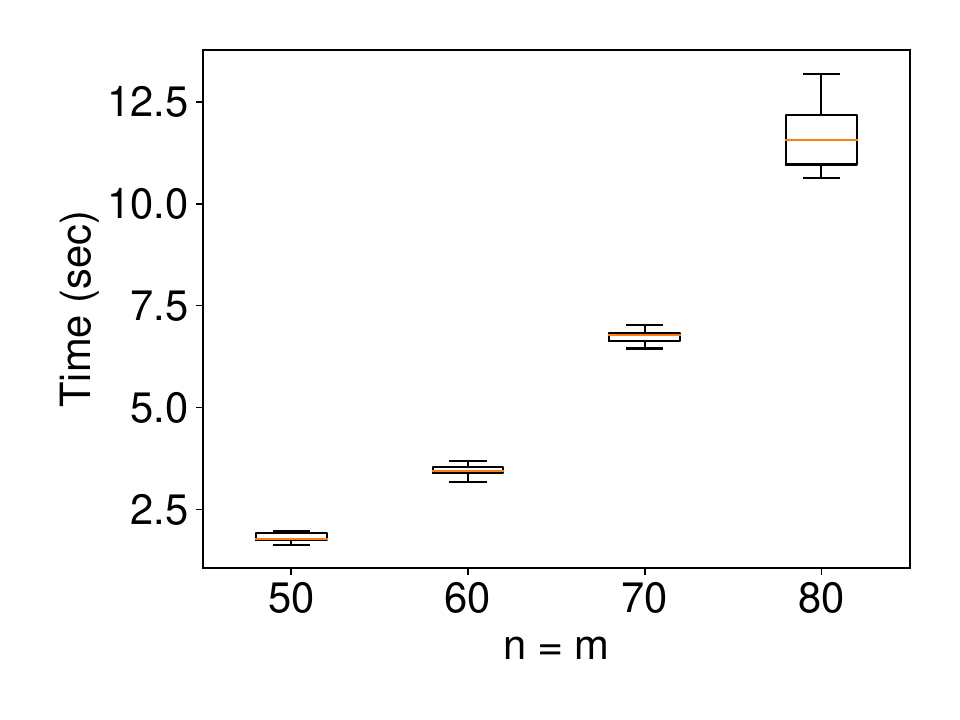}  
  \caption{Computational cost}
\end{subfigure}
\caption{Coverage guarantee, CI length and computational cost in multi-dimensional case ($d = 2$).}
\label{fig:numerical_2_d}
\end{figure}

\subsubsection{Multi-dimensional case ($d = 2$)}
We generated the dataset 
$X = \{\bm x_{i, :} \}_{i \in [n]}$ 
with $\bm x_{i, :} \sim \NN (\bm 1_d, I_d)$
and 
$Y = \{\bm y_{j, :} \}_{j \in [m]}$ 
with $\bm y_{j, :} \sim \NN (\bm 1_d + \Delta, I_d)$  (element-wise addition).
Similar to the univariate case, we set $n = m = 5$ and ran 120 trials for each $\Delta \in \{0, 1, 2, 3, 4\}$ for the experiments of 
coverage guarantee and CI length as well as setting $\Delta = 2$ and ran 10 trials for each $n = m \in \{50, 60, 70, 80\}$ for the experiments of computational cost.
The results are shown in Figure \ref{fig:numerical_2_d}.
The interpretation of the results is similar and consistent with the univariate case.


\subsubsection{Robustness of the proposed selective CI in terms of coverage guarantee}
\begin{figure}[!t]

\begin{subfigure}{.245\linewidth}
  \centering
  \includegraphics[width=\linewidth]{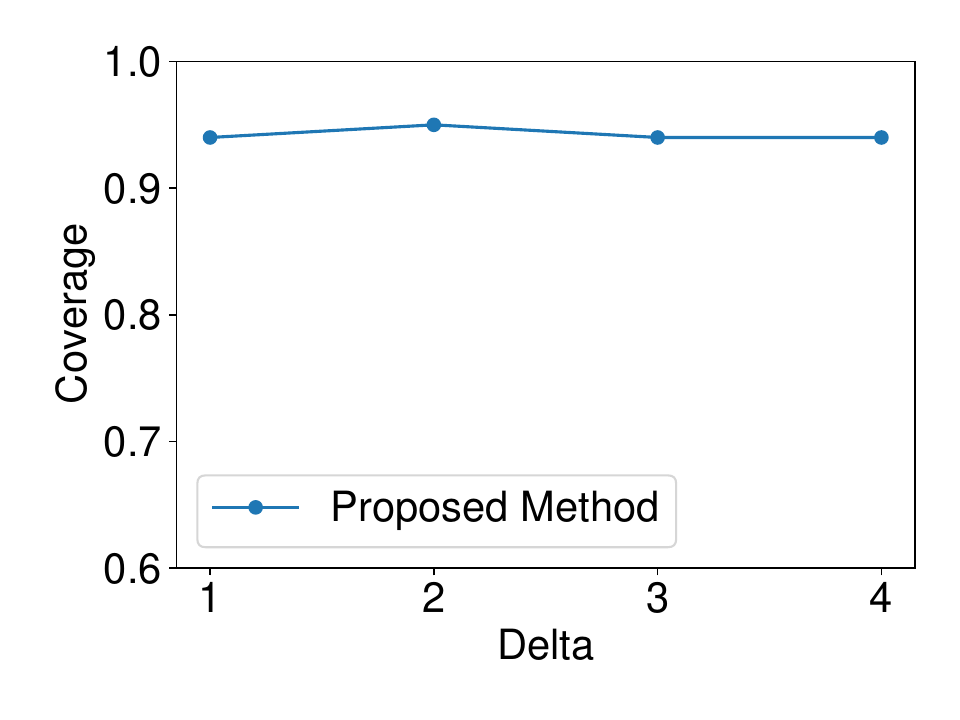}  
  \caption{Laplace}
\end{subfigure}
\begin{subfigure}{.245\linewidth}
  \centering
  \includegraphics[width=\linewidth]{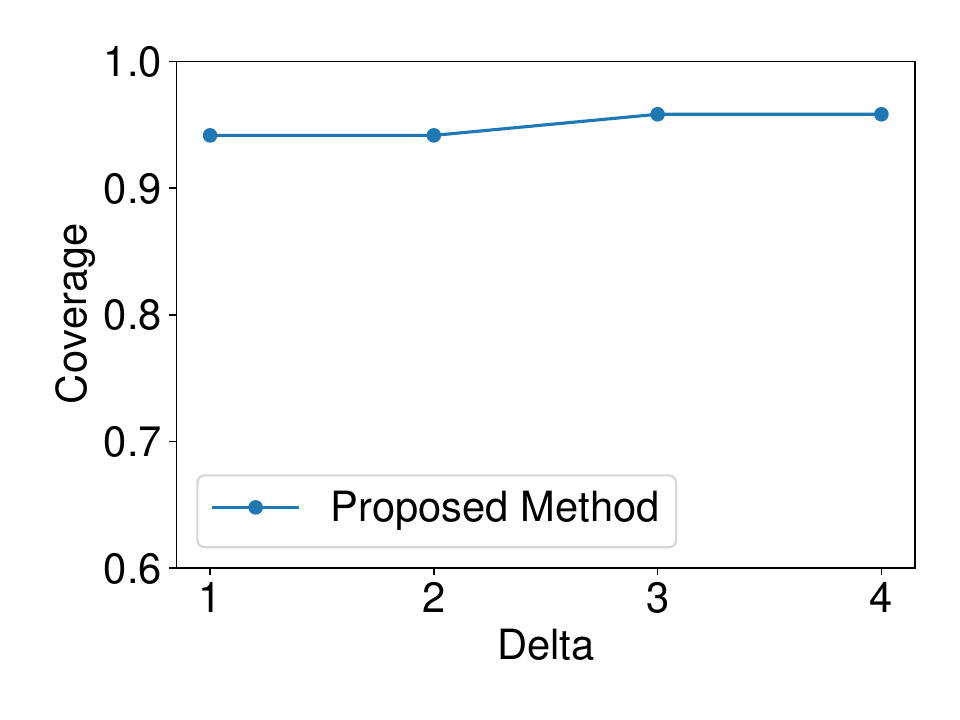}  
  \caption{Skew normal}
\end{subfigure}
\begin{subfigure}{.245\linewidth}
  \centering
  \includegraphics[width=\linewidth]{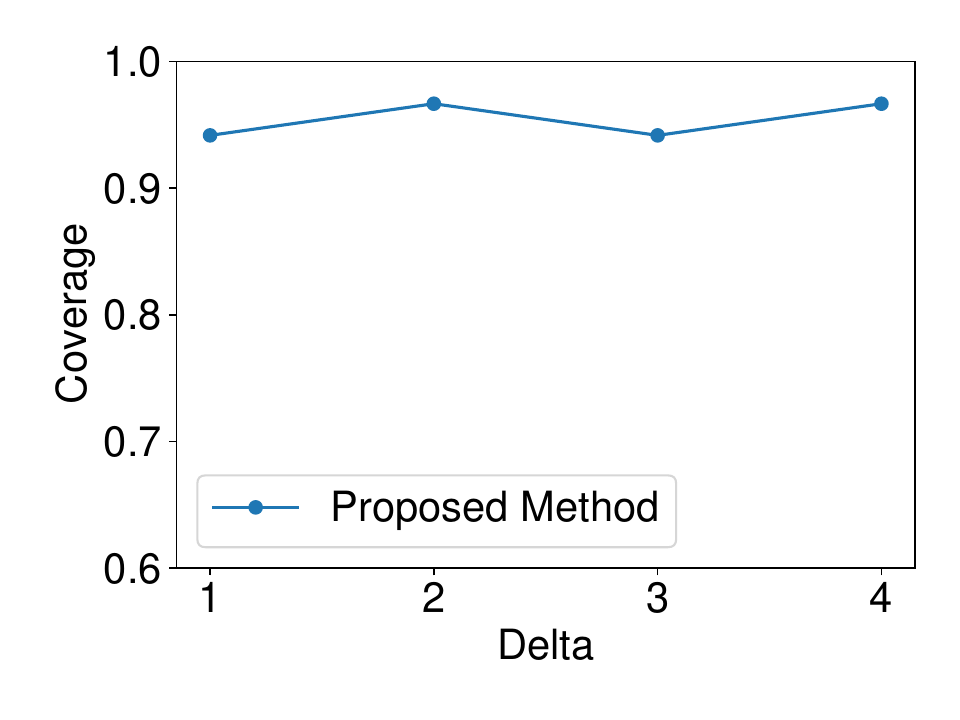}  
  \caption{$t_{20}$}
\end{subfigure}
\begin{subfigure}{.245\linewidth}
  \centering
  \includegraphics[width=\linewidth]{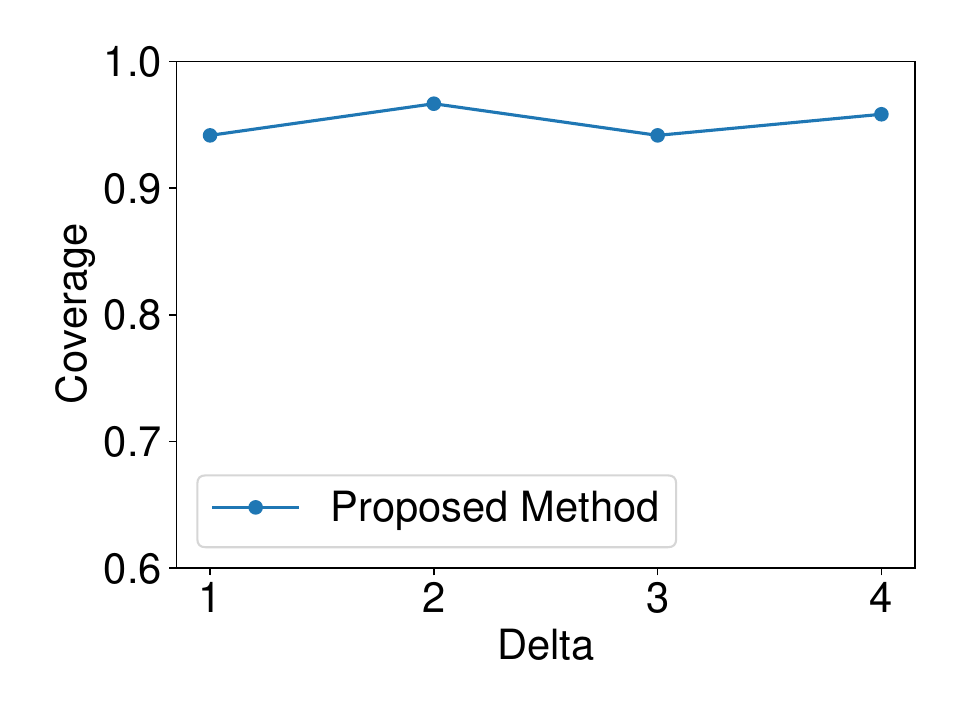}  
  \caption{Estimated variance}
\end{subfigure}
\caption{Robustness of the proposed selective CI in terms of coverage guarantee.}
\label{fig:robust}
\end{figure}

We additionally demonstrate the robustness of the proposed selective CI in terms of coverage guarantee by considering the following cases:
\begin{itemize}
	\item Non-normal noise: we considered the noises $\bm \veps_{\bm X}$ and $\bm \veps_{\bm Y}$ following the Laplace distribution, skew normal distribution (skewness coefficient: $10$), and $t_{20}$ distribution.
	\item Unknown variance: we considered the case in which the variance of the noises was also estimated from the data.
\end{itemize}
The dataset $\bm X$ and $\bm Y$ were generated with 
$\bm \mu_{\bm X} = \bm 1_n$, 
$\bm \mu_{\bm Y} = \bm 1_m + \Delta$.
We set $n = m = 5$ and ran 120 trials for each $\Delta \in \{1, 2, 3, 4\}$.
We confirmed that our selective CI maintained good performance in terms of coverage guarantee.
The results are shown in Figure \ref{fig:robust}.

\subsubsection{Comparison with asymptotic method in \cite{imaizumi2019hypothesis}}
\begin{figure}[!t]

\begin{subfigure}{.245\linewidth}
  \centering
  \includegraphics[width=\linewidth]{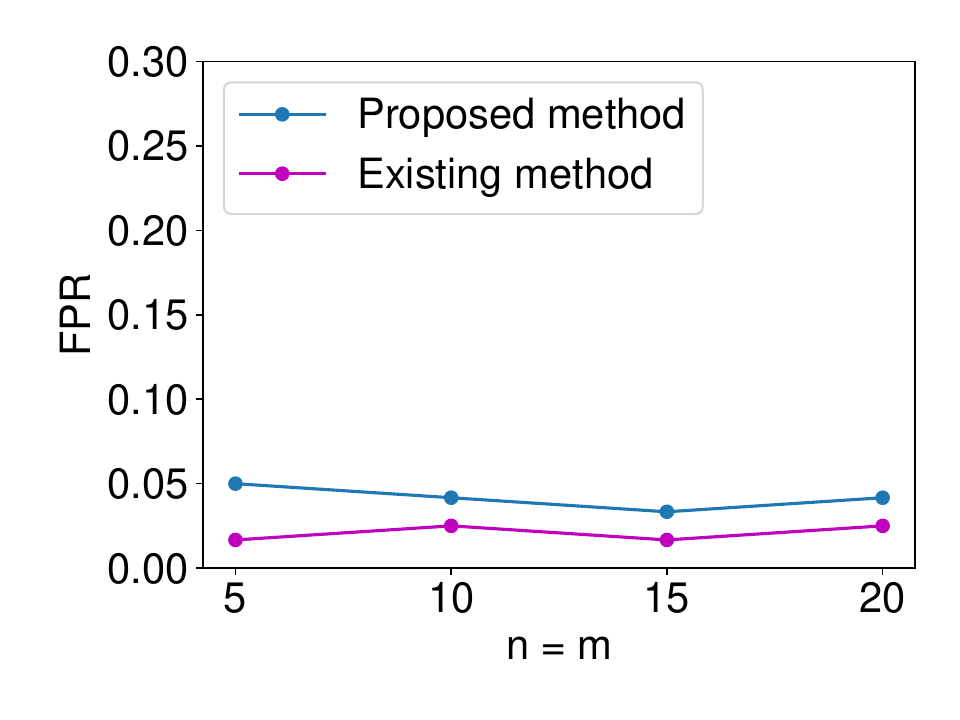}  
  \caption{FPR}
\end{subfigure}
\begin{subfigure}{.245\linewidth}
  \centering
  \includegraphics[width=\linewidth]{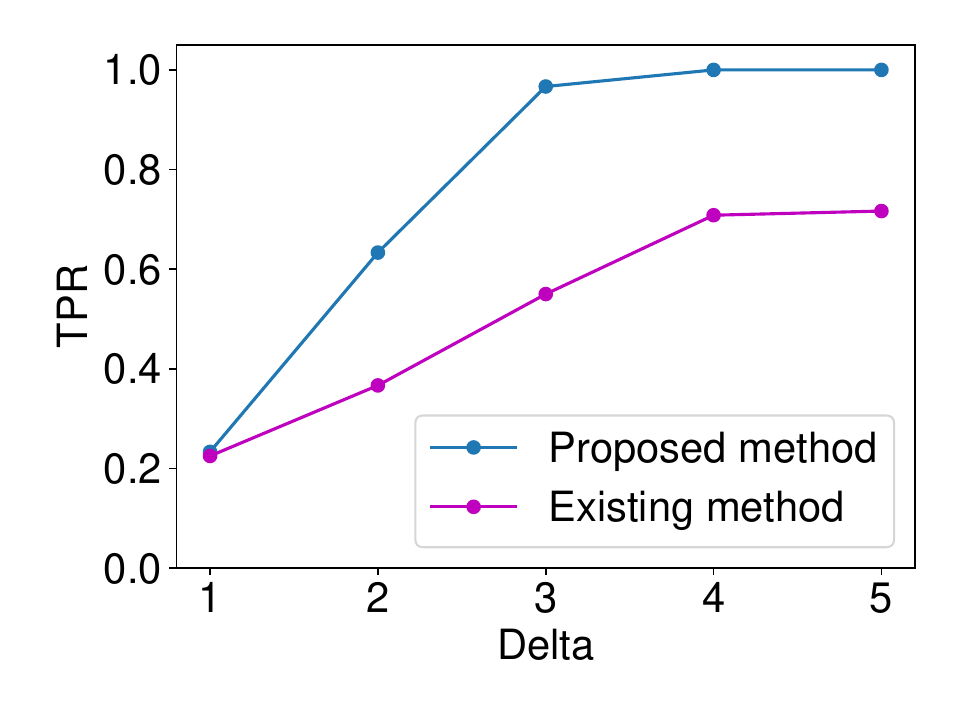}  
  \caption{TPR ($n = m = 5$)}
\end{subfigure}
\begin{subfigure}{.245\linewidth}
  \centering
  \includegraphics[width=\linewidth]{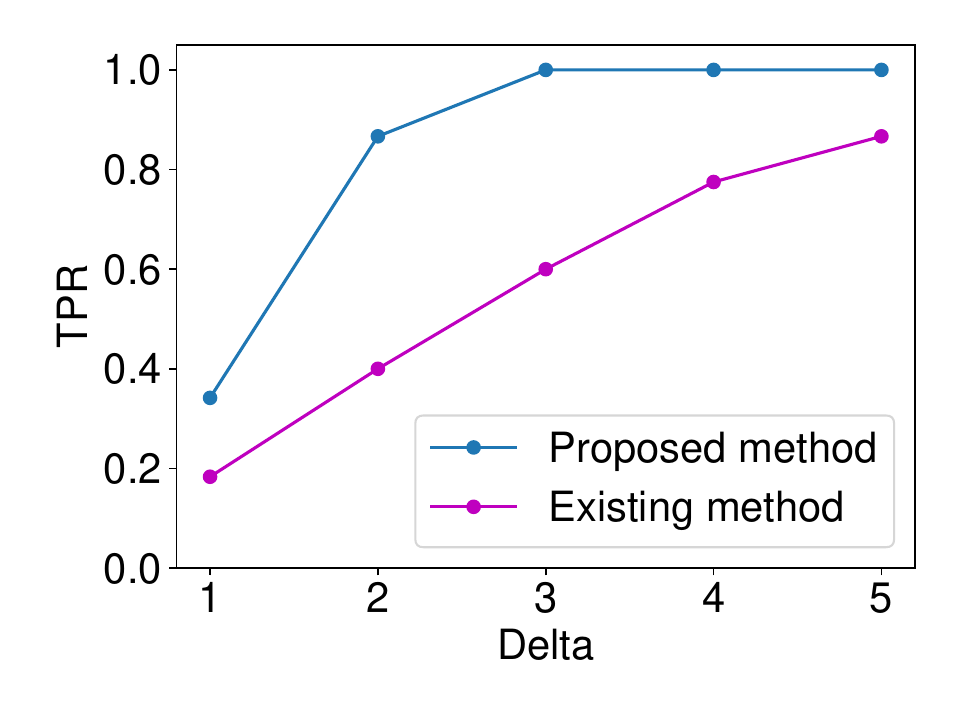}  
  \caption{TPR ($n = m = 10$)}
\end{subfigure}
\begin{subfigure}{.245\linewidth}
  \centering
  \includegraphics[width=\linewidth]{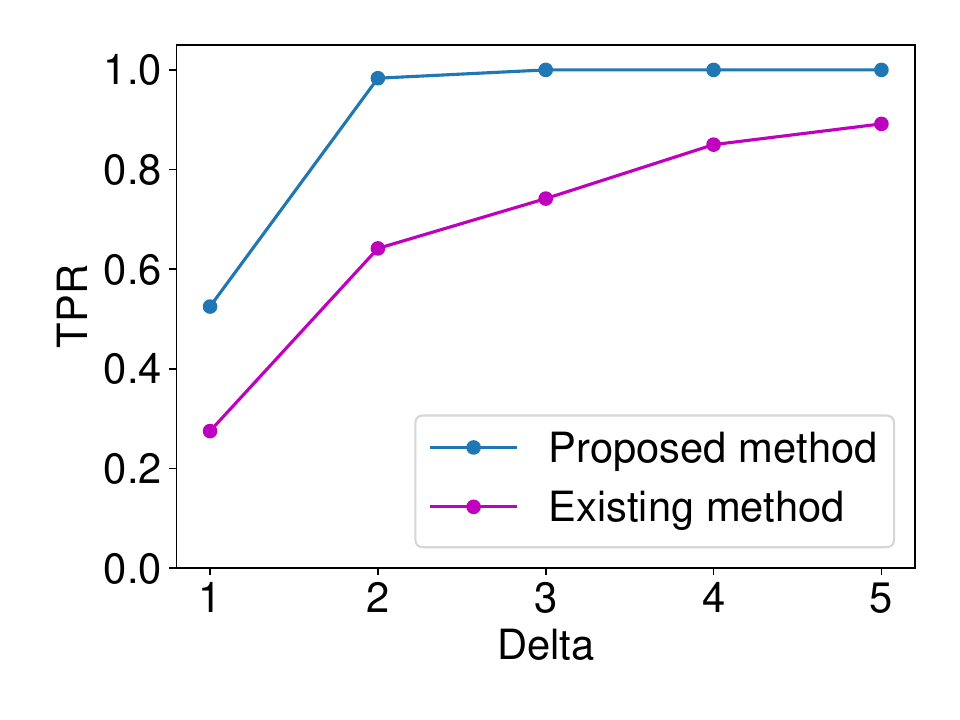}  
  \caption{TPR ($n = m = 20$)}
\end{subfigure}
\caption{Comparisons with asymptotic method \citep{imaizumi2019hypothesis} in terms of FPR and TPR.
While both methods can successfully control the FPR under the significance level $\alpha=0.05$, the proposed method has higher TPR than the existing asymptotic method in all the cases.
}
\label{fig:compare_with_asymptotic_method}
\end{figure}

The authors of \cite{imaizumi2019hypothesis} provided us their code of computing the $p$-value in hypothesis testing framework. Therefore, we conducted the comparisons in terms of false positive rate (FPR) control and true positive rate (TPR).
Although we mainly focused on the CI in the previous sections, the corresponding hypothesis testing problem is defined as follows:
\begin{align*}
	{\rm H}_0: \bm \eta^\top {\bm \mu_{\bm X} \choose \bm \mu_{\bm Y}} = 0 
	\quad 
	\text{v.s.}
	\quad 
	{\rm H}_1: \bm \eta^\top {\bm \mu_{\bm X} \choose \bm \mu_{\bm Y}} \neq 0.
\end{align*}
The details are presented in Appendix \ref{appendix:app_l_2}.
We generated the dataset $\bm X$ and $\bm Y$ with 
$\bm \mu_{\bm X} = \bm 1_n$, 
$\bm \mu_{\bm Y} = \bm 1_m + \Delta$ (element-wise addition),
$\bm \veps_{\bm X} \sim \NN(\bm 0, I_n)$,
$\bm \veps_{\bm Y} \sim \NN(\bm 0, I_m)$.
Regarding the FPR experiments, we set $\Delta = 0$ and and ran 120 trials for each $n = m \in \{ 5, 10, 15, 20\} $.
In regard to the TPR experiments, we set $\Delta \in \{ 1, 2, 3, 4, 5\}$ and ran 120 trials for each $n = m \in \{ 5, 10, 20\} $.
The results are shown in Figure \ref{fig:compare_with_asymptotic_method}.
In terms for FPR control, both methods could successfully control the FPR under $\alpha = 0.05$.
However, in terms of TPR, the proposed method outperformed the existing asymptotic one in all the cases.
As observed in Figure \ref{fig:compare_with_asymptotic_method} (a), the existing method is conservative in the sense that the FPR is smaller than the specified significance level $\alpha$ = 0.05. 
As a consequence of this conservativeness, the power of their method was consistently lower than ours. Such a phenomenon is commonly observed in approximate statistical inference.

\subsection{Real Data Experiment} \label{subsec:real_data}

In this section, we evaluate the proposed selective CI on four real-world datasets.
We used Iris dataset, Wine dataset, Breast Cancer dataset which are available in the UCI machine learning repository, and Lung Cancer dataset \footnote{We used dataset Lung\_GSE7670 which is available at \url{https://sbcb.inf.ufrgs.br/cumida}.} \citep{feltes2019cumida}.
The experiments were conducted with the following settings:
\begin{itemize}
	\item \textbf{Setting 1}: For each pair of classes in the dataset:
	\begin{itemize}
		\item Randomly select $n = m = 5$ instances from each class. 
		Here, each instance is represented by a $d$-dimensional vector where $d$ is the number of features.
		\item Compute the selective CI.
		\item Repeat the above process up to 120 times.
	\end{itemize}
	\item \textbf{Setting 2}: Given a dataset with two classes {\tt C1} and {\tt C2}, we either chose $n = 5$ instances from {\tt C1} and $m = 5$ instances from {\tt C2} ($X^{\rm {obs}}$ and $Y^{\rm {obs}}$ are from different classes); or both $X^{\rm {obs}}$ and $Y^{\rm {obs}}$ from either {\tt C1}  or {\tt C2} ($X^{\rm {obs}}$ and $Y^{\rm {obs}}$ are from the same class). 
	Then, we compute the selective CI.
	We repeated this process up to 120 times.
\end{itemize}


\begin{figure}[!t]

\begin{subfigure}{.33\linewidth}
  \centering
  \includegraphics[width=\linewidth]{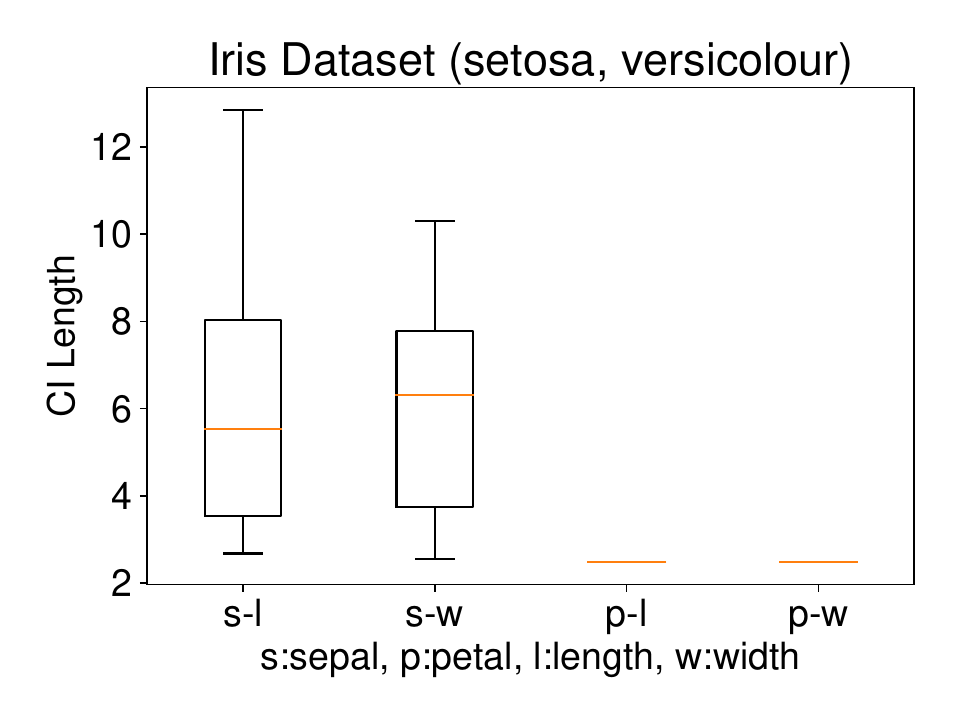}  
  \caption{C1-C2}
\end{subfigure}
\begin{subfigure}{.33\linewidth}
  \centering
  \includegraphics[width=\linewidth]{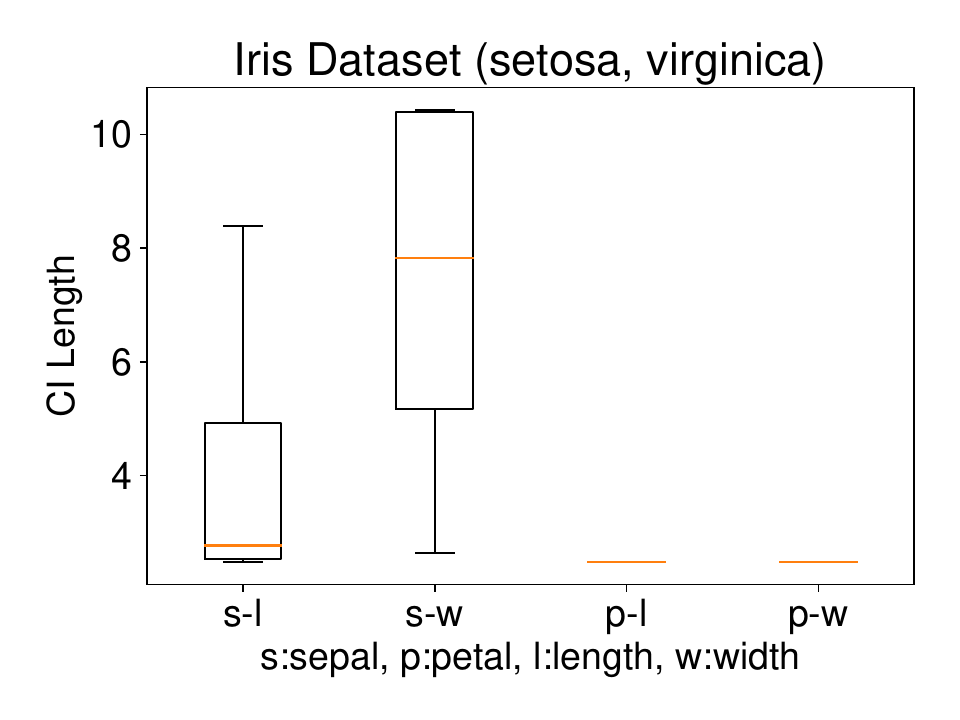}  
  \caption{C1-C3}
\end{subfigure}
\begin{subfigure}{.33\linewidth}
  \centering
  \includegraphics[width=\linewidth]{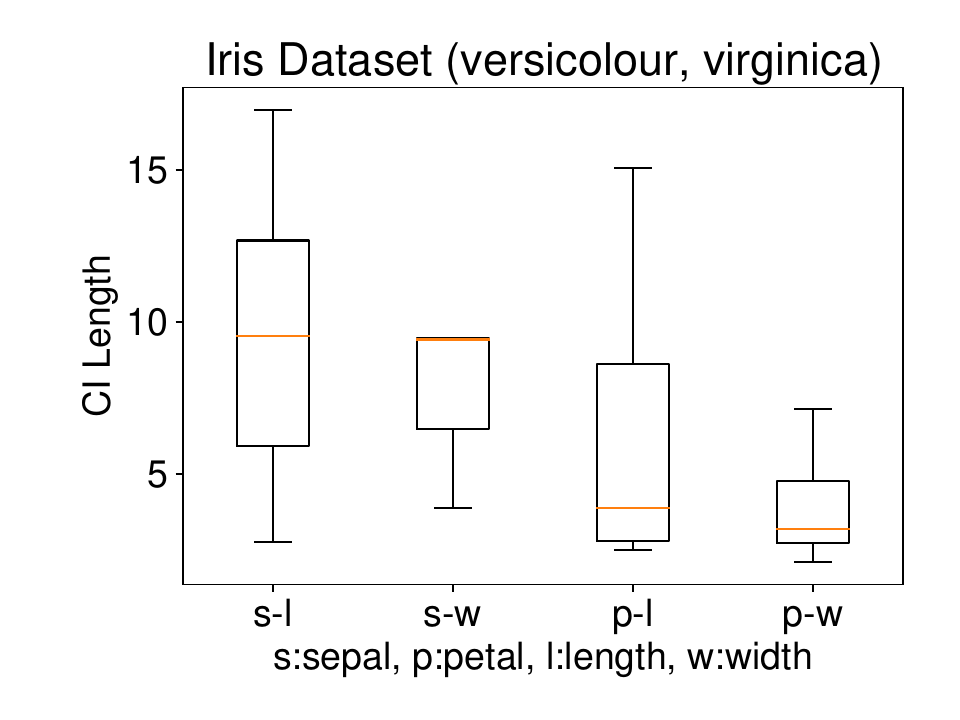}  
  \caption{C2-C3}
\end{subfigure}
\caption{Results on Iris dataset in univariate case ($d = 1$).}
\label{fig:real_iris_1_d}
\end{figure}

\subsubsection{Univariate case ($d = 1$) with Setting 1}
We conducted the experiments on Iris dataset which contains three classes: Iris Setosa (C1), Iris Versicolour (C2), and Iris Virginica (C3).
This dataset also contains four features: sepal length ({\tt s-l}), sepal width ({\tt s-w}), petal length ({\tt p-l}), and petal width ({\tt p-w}).
We ran the procedure described in Setting 1 on each individual feature.
The results are shown in Figure \ref{fig:real_iris_1_d}.
In all three plots of this figure, the two features {\tt p-l} and {\tt p-w }always have the shortest CI length among the four features which indicates that these two features are informative to discriminate between the classes.
Besides, the results of Figure \ref{fig:real_iris_1_d} are also consistent with the results obtained after plotting the histogram of each feature in each class.
In other words, the farther the two histograms, the smaller the length of selective CI.


\begin{figure}[!t]

\begin{subfigure}{.495\linewidth}
  \centering
  	\begin{minipage}{.49\linewidth}
		\centering
		\includegraphics[width=\textwidth]{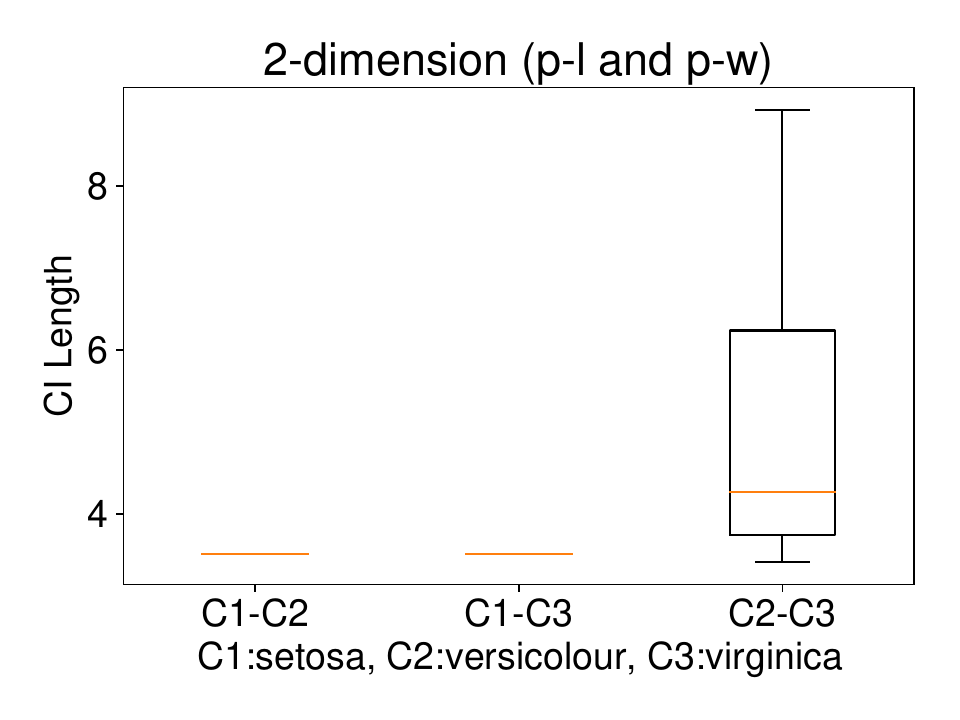}
	\end{minipage}
	\begin{minipage}{.49\linewidth}
		\centering
		\includegraphics[width=\textwidth]{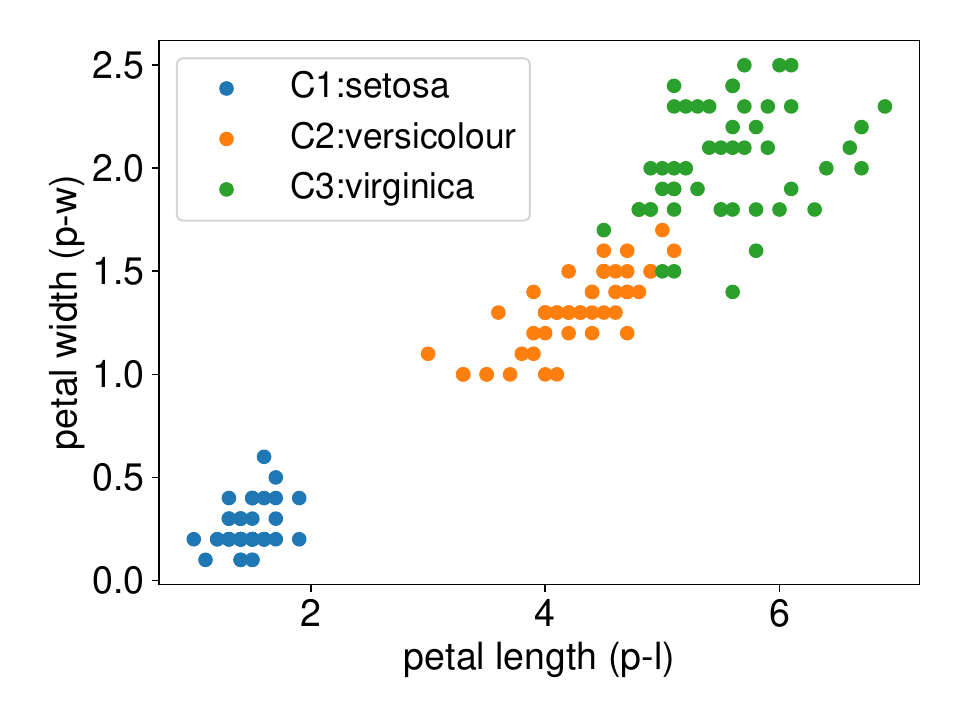}
	\end{minipage} 
	\caption{$d = 2$}
\end{subfigure}
\begin{subfigure}{.495\linewidth}
  \centering
  	\begin{minipage}{.49\linewidth}
		\centering
		\includegraphics[width=\textwidth]{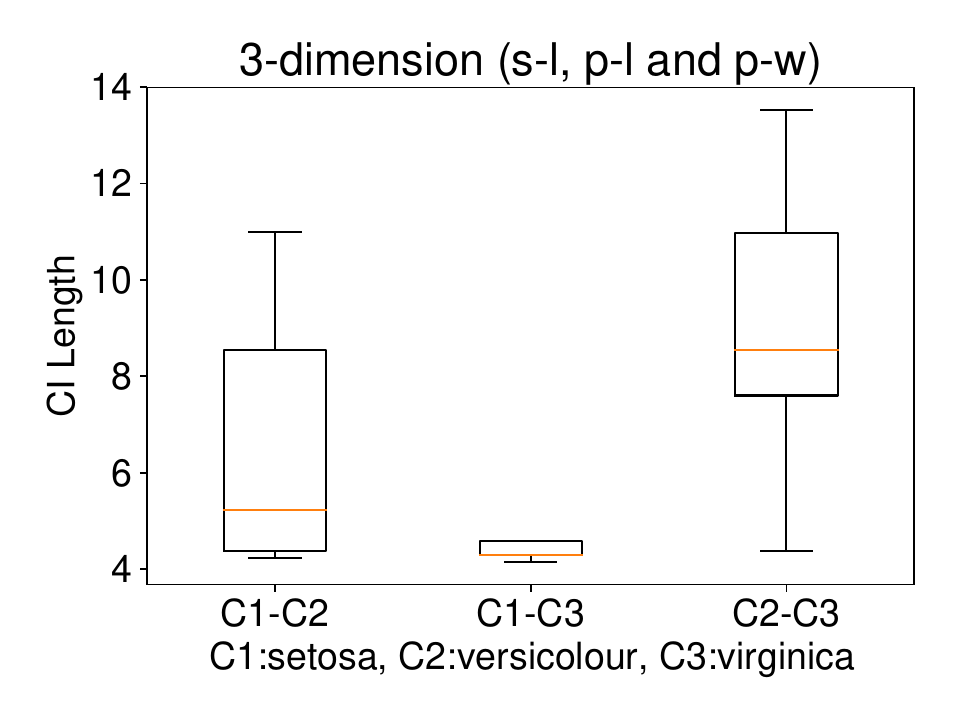}
	\end{minipage}
	\begin{minipage}{.49\linewidth}
		\centering
		\includegraphics[width=\textwidth]{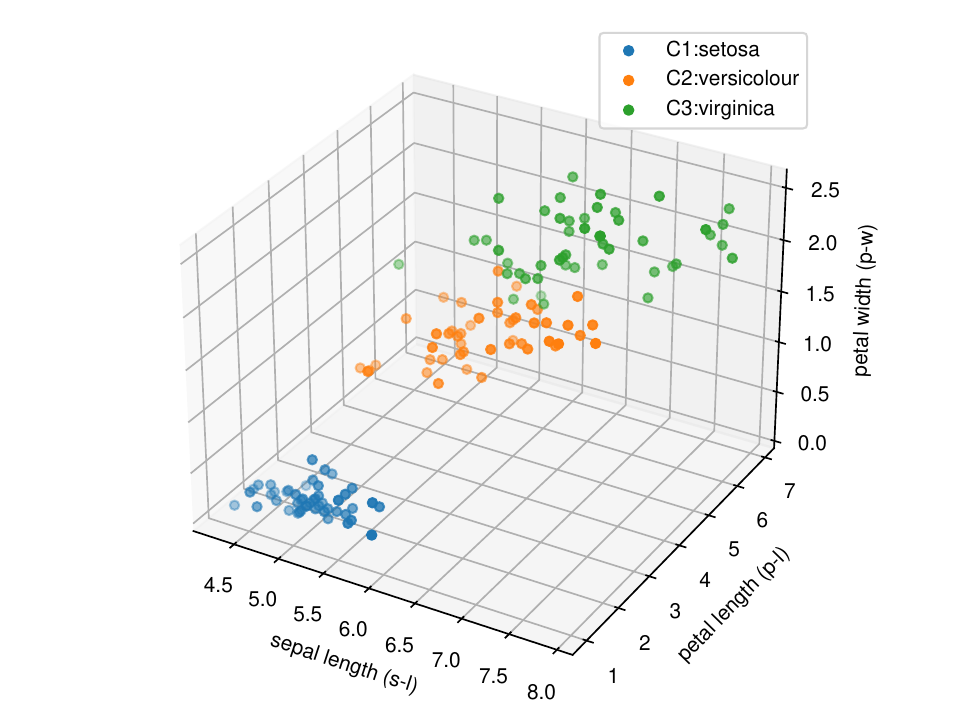}
	\end{minipage} 
  \caption{$d = 3$}
\end{subfigure}
\caption{Results on Iris dataset in multi-dimensional case ($d \in \{ 2, 3 \} $)}
\label{fig:real_iris_2_3_d}
\end{figure}

\subsubsection{Multi-dimensional case ($d \in \{2, 3 \} $) with Setting 1}
\begin{figure}[!t]

\begin{subfigure}{.495\linewidth}
  \centering
  	\begin{minipage}{.49\linewidth}
		\centering
		\includegraphics[width=\textwidth]{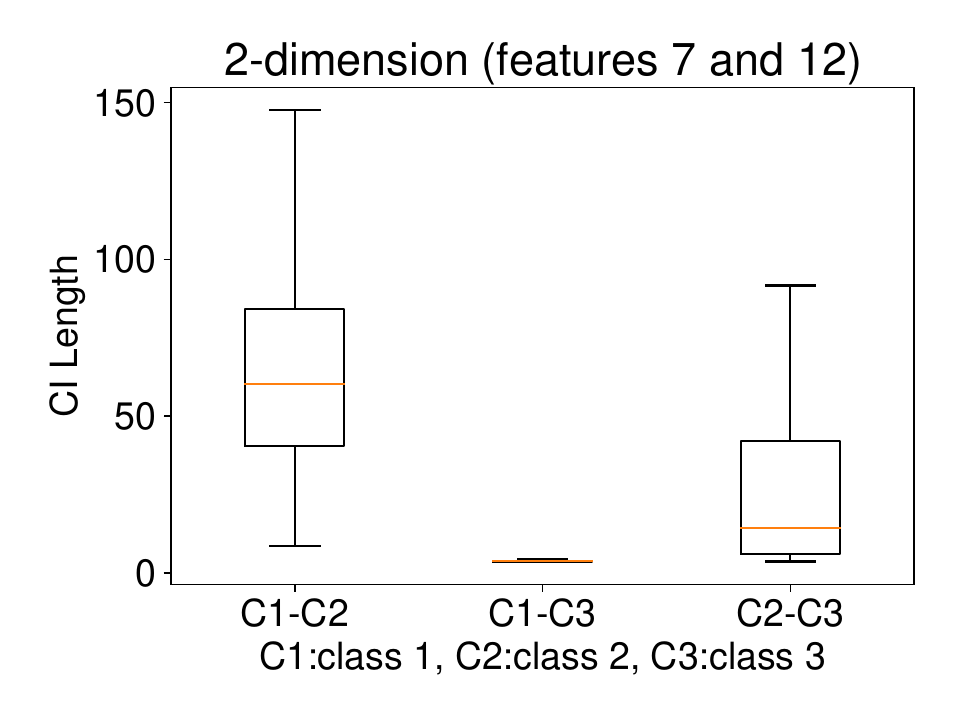}
	\end{minipage}
	\begin{minipage}{.49\linewidth}
		\centering
		\includegraphics[width=\textwidth]{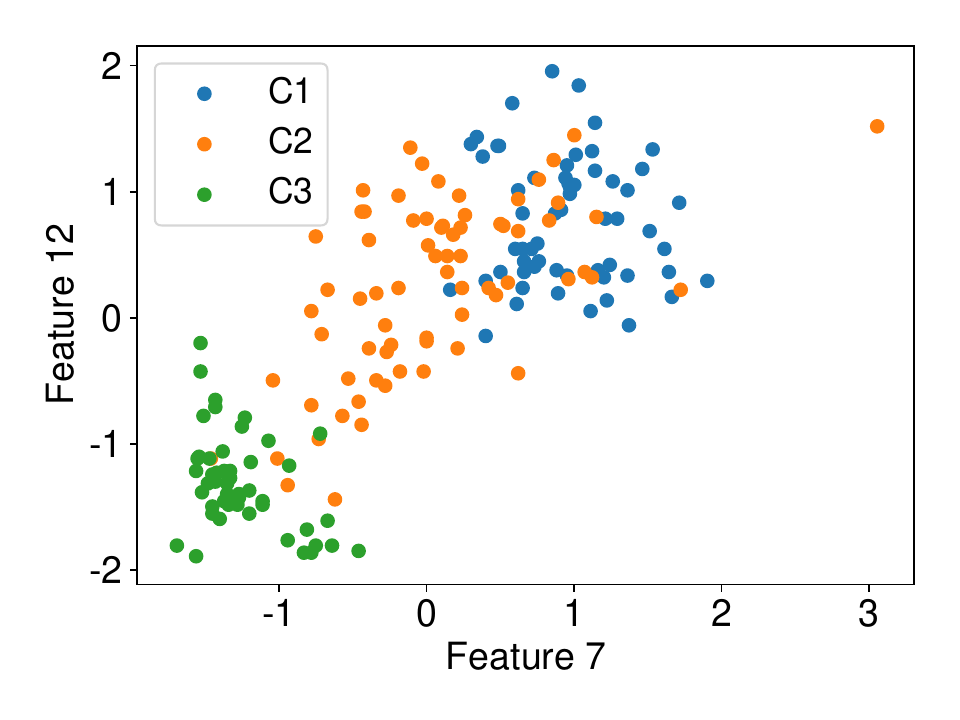}
	\end{minipage} 
	\caption{$d = 2$ (features 7 and 12)}
\end{subfigure}
\begin{subfigure}{.495\linewidth}
  \centering
  	\begin{minipage}{.49\linewidth}
		\centering
		\includegraphics[width=\textwidth]{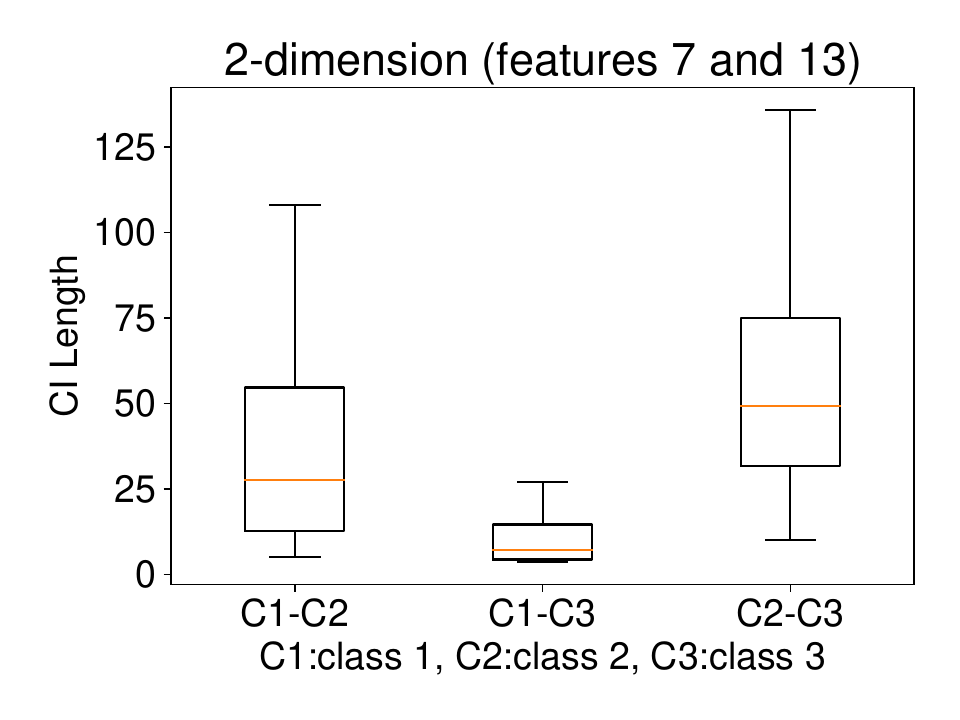}
	\end{minipage}
	\begin{minipage}{.49\linewidth}
		\centering
		\includegraphics[width=\textwidth]{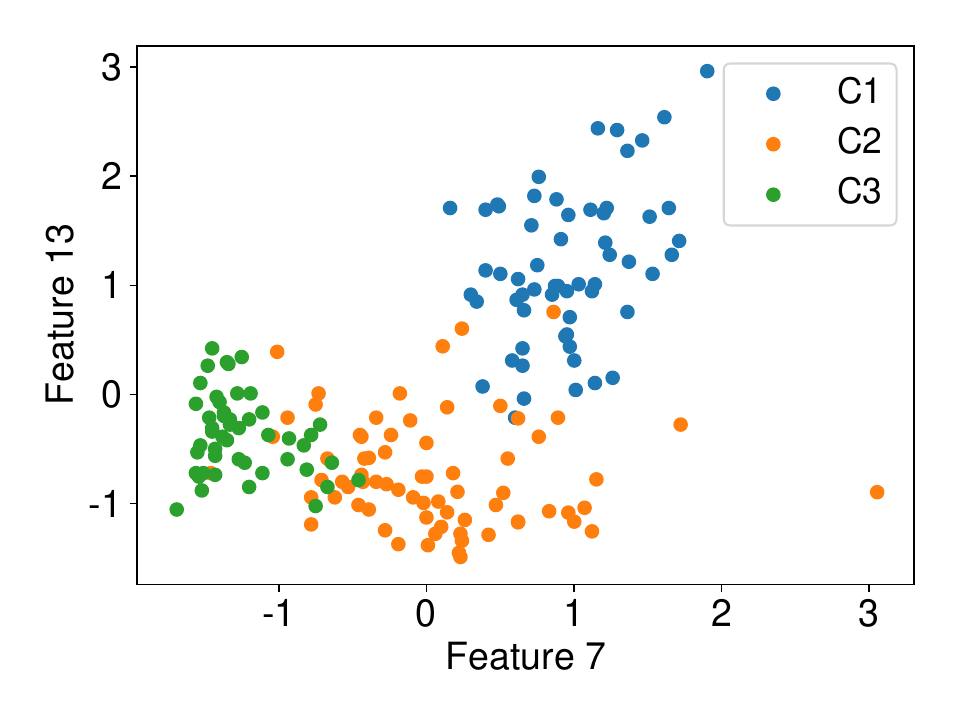}
	\end{minipage} 
  \caption{$d = 2$ (features 7 and 13)}
\end{subfigure}
\begin{subfigure}{.495\linewidth}
  \centering
  	\begin{minipage}{.49\linewidth}
		\centering
		\includegraphics[width=\textwidth]{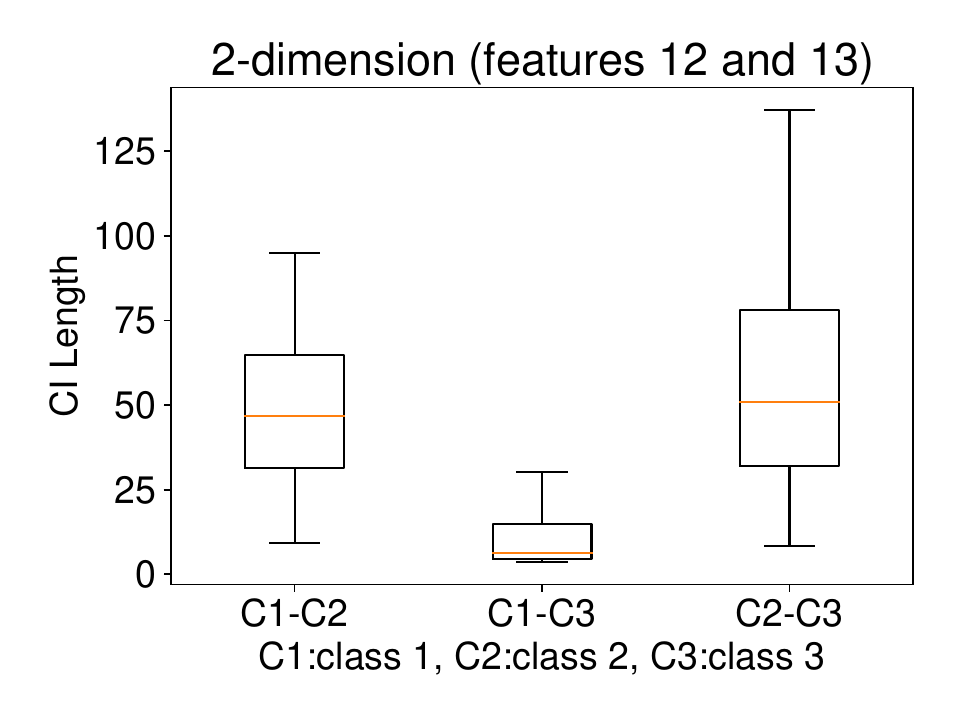}
	\end{minipage}
	\begin{minipage}{.49\linewidth}
		\centering
		\includegraphics[width=\textwidth]{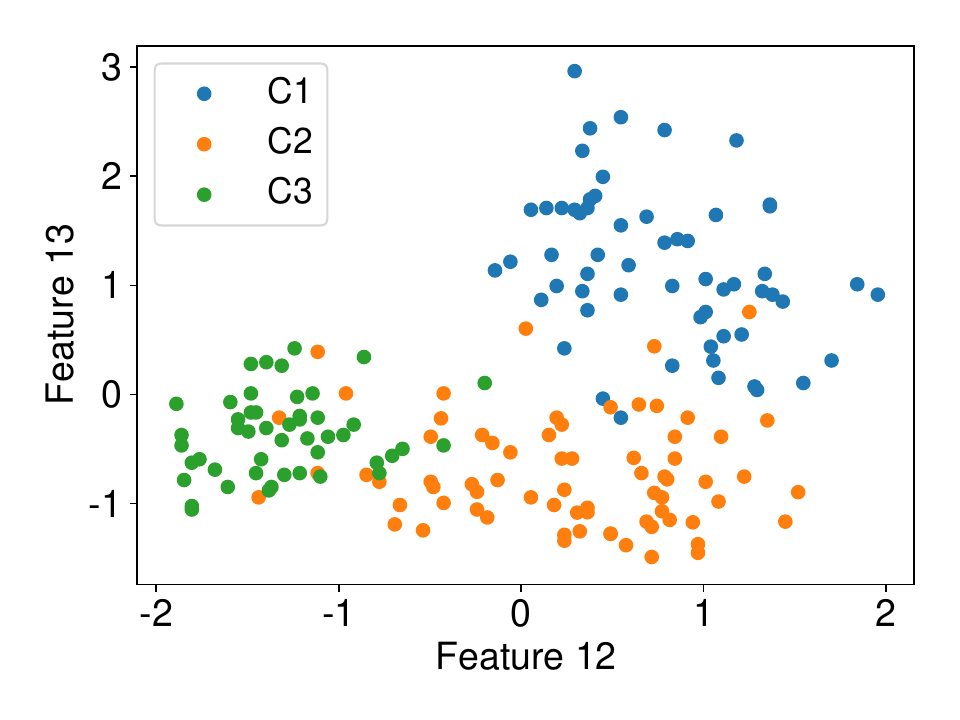}
	\end{minipage} 
	\caption{$d = 2$ (features 12 and 13)}
\end{subfigure}
\begin{subfigure}{.495\linewidth}
  \centering
  	\begin{minipage}{.49\linewidth}
		\centering
		\includegraphics[width=\textwidth]{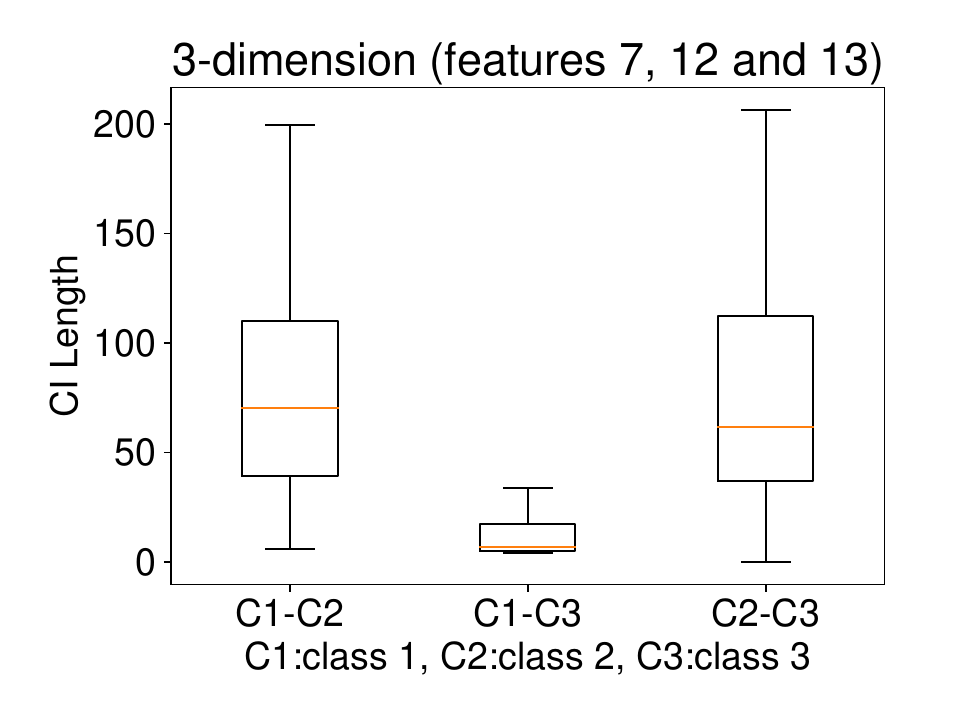}
	\end{minipage}
	\begin{minipage}{.49\linewidth}
		\centering
		\includegraphics[width=\textwidth]{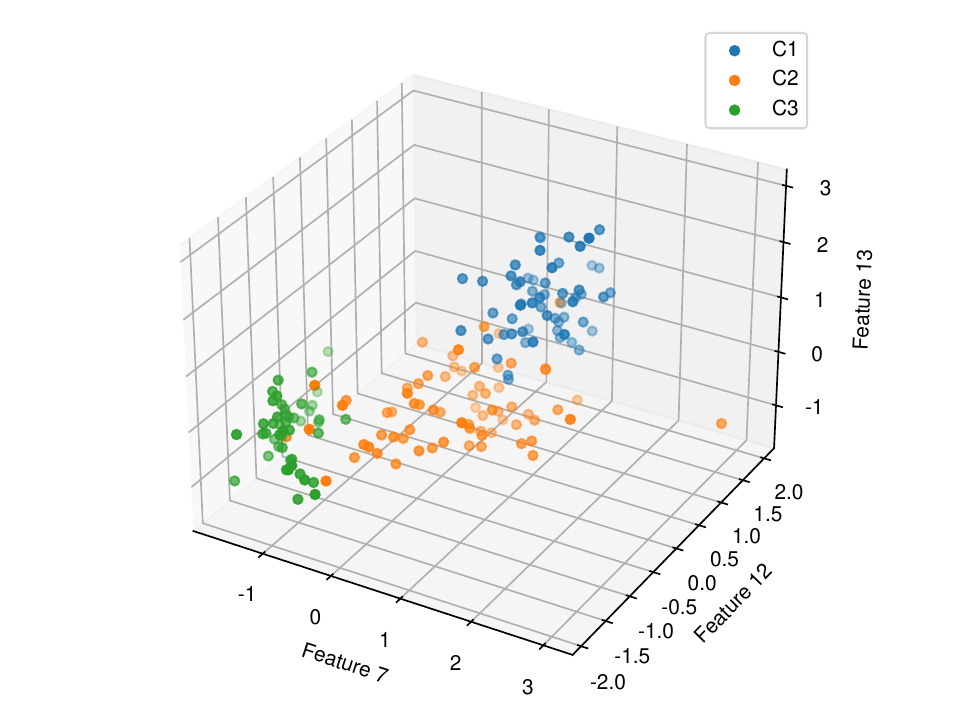}
	\end{minipage} 
  \caption{$d = 3$ (features 7, 12, and 13)}
\end{subfigure}
\caption{Results on Wine dataset in multi-dimensional case ($d \in \{ 2, 3 \} $)}
\label{fig:real_wine_2_3_d}
\end{figure}

Regarding the experiments on Iris dataset in multi-dimensional case, we chose two features {\tt p-l} and {\tt p-w} when $d = 2$ and additionally include feature {\tt s-l} when $d = 3$.
The results are shown in Figure \ref{fig:real_iris_2_3_d}.
In each sub-plot, we show the results of the length of selective CI and the corresponding scatter plot which is used to verify the CI length results.
For example, in Figure \ref{fig:real_iris_2_3_d}a, it is obvious that the distance between {\tt C1} and {\tt C2} as well as the distance between {\tt C1} and {\tt C3} are larger than the distance between {\tt C2} and {\tt C3} by seeing the scatter plot. 
Therefore, the CI lengths of {\tt C1-C2} and {\tt C1-C3} tend to be smaller than that of {\tt C2-C3}.
Besides, we also additionally conducted experiments on Wine dataset.
This dataset contains 3 classes of wine and 13 features.
In the case of $d=2$, we conducted the experiments on each pair features in the set $\{7, 12, 13\}$ (feature 7: flavanoids, features 12: od280/od315 of diluted wines, feature 13: proline).
In the case of $d = 3$, we conducted the experiments on both three features.
The results are shown in Figure \ref{fig:real_wine_2_3_d}.
In general, the results of CI length are consistent with the scatter plots, i.e., the farther the scatter plots between two classes, the smaller the length of selective CI.


\subsubsection{Multi-dimensional case with Setting 2} 
\begin{figure}[!t]

\begin{subfigure}{.48\linewidth}
  \centering
  \includegraphics[width=.8\linewidth]{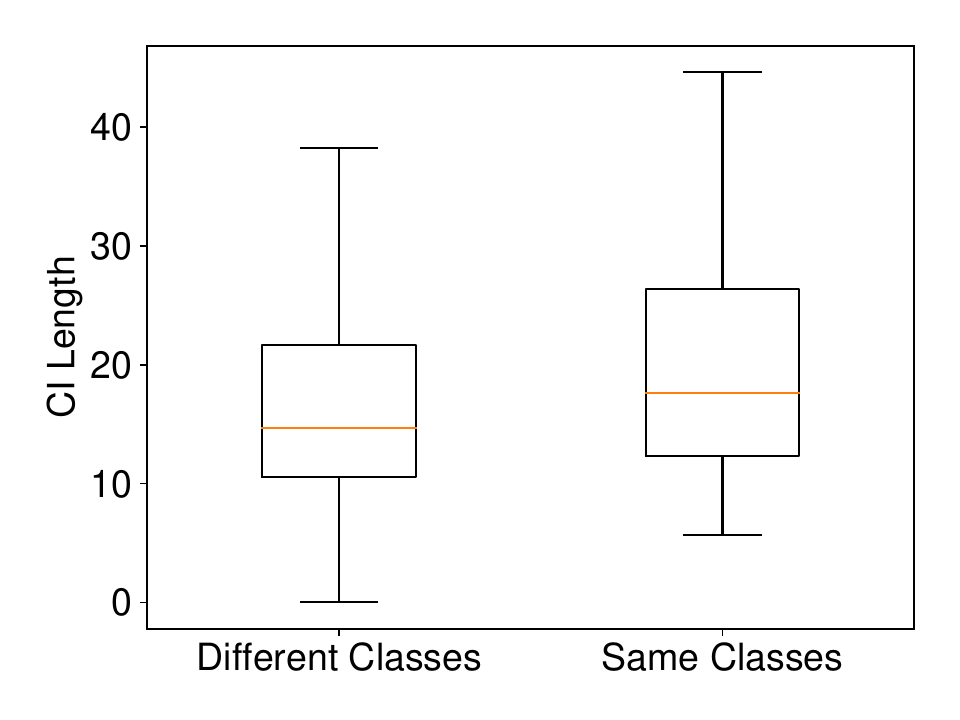}  
  \caption{Breast Cancer dataset ($d = 30$)}
\end{subfigure}
\hspace{2pt}
\begin{subfigure}{.48\linewidth}
  \centering
  \includegraphics[width=.8\linewidth]{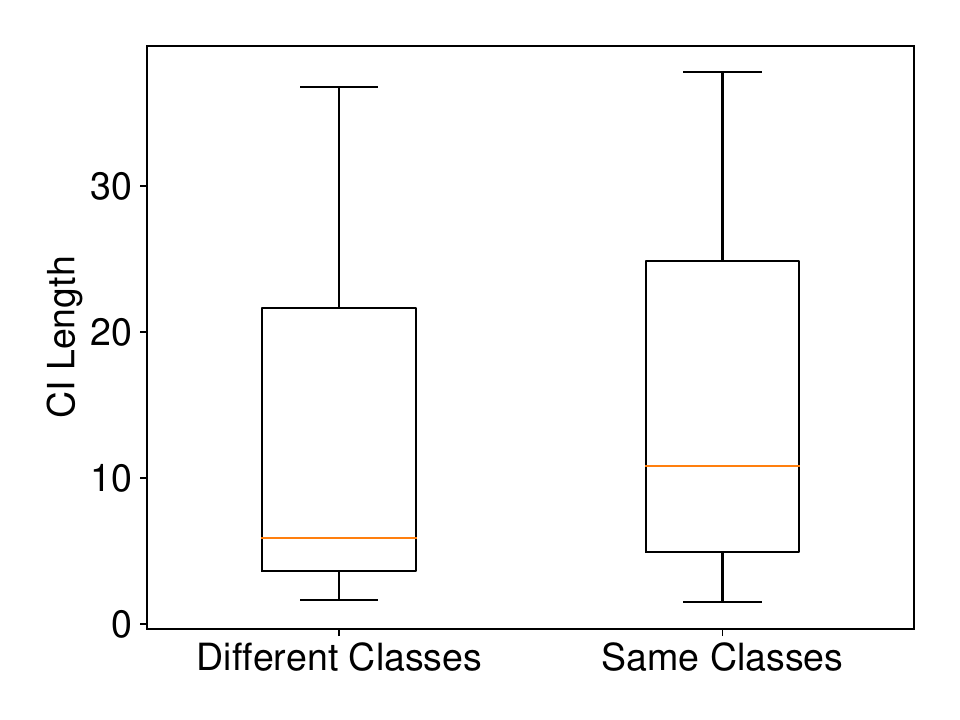}  
  \caption{Lung Cancer dataset ($d = 1,000$)}
\end{subfigure}
\caption{Results on Breast Cancer and Lung Cancer datasets in multi-dimensional case.}
\label{fig:real_breast_lung}
\end{figure}

We conducted experiments on Breast Cancer and Lung Cancer datasets. 
In Breast Cancer dataset, there are two classes (malignant and benign) and $d = 30$ features.
In Lung Cancer dataset, there are two classes (normal and adenocarcinoma) and we choose $d = 1,000$ (we selected the top 1,000 genes which have the largest standard deviations as it is commonly done in the literature).
The results on these datasets with Setting 2 are shown in Figure \ref{fig:real_breast_lung}.
The results are consistent with the intuitive expectation.
When $X^{\rm {obs}}$ and $Y^{\rm {obs}}$ are from different classes, the Wasserstein distance tends to be larger than the one computed when $X^{\rm {obs}}$ and $Y^{\rm {obs}}$ are from the same class.
Therefore, the CI for the Wasserstein distance in the case of different classes is shorter than the one computed in the case of same class.
In other words, the larger the Wasserstein distance is, the shorter the CI becomes.

\section{Conclusion} \label{sec:conclusion}
In this paper, we present an exact (non-asymptotic) statistical inference method for the Wasserstein distance.
We first introduce the problem setup and present the proposed method for univariate case.
We next provide the extension to multi-dimensional case.
We finally conduct the experiments on both synthetic and real-world datasets to evaluate the performance of our method.
To our knowledge, this is the first method that can provide a valid confidence interval (CI) for the Wasserstein distance with finite-sample coverage guarantee.
We believe this study is an important contribution toward reliable ML, which is one of the most critical issues in the ML community.





\bibliographystyle{abbrvnat}
\bibliography{ref}


\appendix
\section{Proposed method in hypothesis testing framework}\label{appendix:app_l_2}

We present the proposed method in the setting of hypothesis testing and consider the case when the cost matrix is defined by using squared $\ell_2$ distance.

\textbf{Cost matrix.} We define the cost matrix $C(\bm X, \bm Y)$ of pairwise distances (squared $\ell_2$ distance) between elements of $\bm X$ and $\bm Y$ as 
\begin{align} \label{eq:cost_matrix_l_2}
	C(\bm X, \bm Y) 
	& = \big[(x_i - y_j)^2 \big]_{ij} \in \RR^{n \times m}.
\end{align}
Then, the vectorized form of  $C(\bm X, \bm Y)$ can be defined as 
\begin{align} \label{eq:cost_matrix_l_2_vec}
\begin{aligned}
	\bm c(\bm X, \bm Y) 
	&= {\rm {vec}} (C(\bm X, \bm Y)) \in \RR^{nm}\\ 
	&= \left [ \Omega {\bm X \choose \bm Y} \right] \circ \left [ \Omega {\bm X \choose \bm Y} \right],
\end{aligned}
\end{align}
where $\Omega$ is defined as in \eq{eq:matrix_Theta}  and the operation $\circ$ is element-wise product.

\textbf{The Wasserstein distance.} By solving LP with the cost vector defined in \eq{eq:cost_matrix_l_2_vec} on the observed data $\bm X^{\rm {obs}}$ and  $\bm Y^{\rm {obs}}$, we obtain the set of selected basic variables
\begin{align*}
	\cM_{\rm {obs}} = \cM(\bm X^{\rm {obs}}, \bm Y^{\rm {obs}}).
\end{align*}
Then, the Wasserstein distance can be re-written as (we denote $W = W(P_n, Q_m)$ for notational simplicity)
\begin{align*}
	W&= \hat{\bm t}^\top  \bm c(\bm X^{\rm {obs}}, \bm Y^{\rm {obs}}) \\ 
	& = \hat{\bm t}_{\cM_{\rm {obs}}}^\top \bm c_{\cM_{\rm {obs}}}(\bm X^{\rm {obs}}, \bm Y^{\rm {obs}}) \\ 
	& =  \hat{\bm t}_{\cM_{\rm {obs}}}^\top 
	\Bigg [ 
	\left [ \Omega_{\cM_{\rm {obs}}, :} {\bm X^{\rm {obs}} \choose \bm Y^{\rm {obs}}} \right] \circ \left [ \Omega_{\cM_{\rm {obs}}, :} {\bm X^{\rm {obs}} \choose \bm Y^{\rm {obs}}} \right]
	\Bigg ].
\end{align*}

\textbf{Hypothesis testing.} Our goal is to test the following hypothesis:
\begin{align*}
	{\rm H}_0:
	\hat{\bm t}_{\cM_{\rm {obs}}}^\top 
	\Bigg [ 
	\left [ \Omega_{\cM_{\rm {obs}}, :} { \bm \mu_{\bm X} \choose \bm \mu_{\bm Y}} \right] \circ \left [ \Omega_{\cM_{\rm {obs}}, :} { \bm \mu_{\bm X} \choose \bm \mu_{\bm Y}} \right]
	\Bigg ] = 0.
\end{align*}
Unfortunately, it is technically difficult to directly test the above hypothesis.
Therefore, we propose to test the following equivalent one:
\begin{align*}
	& {\rm H}_0:
	\hat{\bm t}_{\cM_{\rm {obs}}}^\top
	\left [ \Theta_{\cM_{\rm {obs}}, :} { \bm \mu_{\bm X} \choose \bm \mu_{\bm Y}} \right]  = 0 \\ 
	\Leftrightarrow ~ &{\rm H}_0: 
	\bm \eta^\top { \bm \mu_{\bm X} \choose \bm \mu_{\bm Y}} = 0
\end{align*}
where $\Theta$ is defined as in \eq{eq:matrix_Theta} and $\bm \eta = \Theta_{\cM_{\rm {obs}}, :}^\top \hat{\bm t}_{\cM_{\rm {obs}}}$.

\textbf{Conditional SI.} To test the aforementioned hypothesis, we consider the following selective $p$-value:
\begin{align*}
	p_{\rm {selective}} = \PP_{\rm H_0} 
	\left (
	\left |\bm \eta^\top {\bm X \choose \bm Y} \right | 
	\geq 
	\left |\bm \eta^\top {\bm X^{\rm {obs}} \choose \bm Y^{\rm {obs}}} \right |
	\mid \cE
	\right )
\end{align*}
where the conditional selection event is defined as 
\begin{align*}
	\cE = 
	\left \{ 
	\begin{array}{l}
	\cM(\bm X, \bm Y) = \cM_{\rm {obs}},  \\ 
	\cS_{\cM_{\rm {obs}}}(\bm X, \bm Y) = \cS_{\cM_{\rm {obs}}}(\bm X^{\rm {obs}}, \bm Y^{\rm {obs}}) \\ 
	\bm q(\bm X, \bm Y) = \bm q(\bm X^{\rm {obs}}, \bm Y^{\rm {obs}})
	\end{array}
	\right \}.
\end{align*}
Our next task is to identify the conditional data space whose data satisfies $\cE$.

\textbf{Characterization of the conditional data space.}
Similar to the discussion in \S \ref{sec:proposed_method}, the data is restricted on the line due to the conditioning on the nuisance component $\bm q(\bm X, \bm Y)$.
Then, the conditional data space is defined as 
\begin{align*} 
	\cD = \Big \{ (\bm X ~ \bm Y)^\top = \bm a + \bm b z \mid z \in \cZ \Big \},
\end{align*}
where 
\begin{align*}
	\cZ = \left \{ 
	z \in \RR ~
	\Big |  
	\begin{array}{l}
	\cM(\bm a + \bm b z) = \cM_{\rm {obs}}, \\ 
	\cS_{\cM_{\rm {obs}}}(\bm a + \bm b z) = \cS_{\cM_{\rm {obs}}}(\bm X^{\rm {obs}}, \bm Y^{\rm {obs}})
	\end{array}
	\right \}.
\end{align*}
The remaining task is to construct $\cZ$.
We can decompose $\cZ$ into two separate sets as 
$\cZ = \cZ_1 \cap \cZ_2$,  where 
\begin{align*}
	\cZ_1 &= \{ z \in \RR \mid  \cM(\bm a + \bm b z) = \cM_{\rm {obs}}\}, \\ 
	\cZ_2 &= \{ z \in \RR \mid \cS_{\cM_{\rm {obs}}}(\bm a + \bm b z) = \cS_{\cM_{\rm {obs}}}(\bm X^{\rm {obs}}, \bm Y^{\rm {obs}}) \}.
\end{align*}
The construction of $\cZ_2$ is as follows (we denote $\bm s_{\cM_{\rm {obs}}} = \cS_{\cM_{\rm {obs}}}(\bm X^{\rm {obs}}, \bm Y^{\rm {obs}})$ for notational simplicity):
\begin{align*}	
	\cZ_2 
	&= \{ z \in \RR \mid \cS_{\cM_{\rm {obs}}}(\bm a + \bm b z) =  \bm s_{\cM_{\rm {obs}}}\} \\ 
	& = \left \{ z \in \RR \mid {\rm {sign}} \Big (\Omega_{\cM_{\rm {obs}}, :} (\bm a + \bm b z) \Big ) =  \bm s_{\cM_{\rm {obs}}} \right \} \\ 
	& = \left \{ z \in \RR \mid \bm s_{\cM_{\rm {obs}}} \circ \Omega_{\cM_{\rm {obs}}, :} (\bm a + \bm b z)  \geq \bm 0\right \},
\end{align*}
which can be obtained by solving the system of linear inequalities.
Next, we present the identification of $\cZ_1$.
Because we use squared $\ell_2$ distance to define the cost matrix, the LP with the parametrized data $\bm a + \bm b z$ is written as follows:
\begin{align*} 
	& \min \limits_{\bm t \in \RR^{nm}} ~  
	\bm t^\top 
	[ \Omega (\bm a + \bm b z) 
	\circ
	\Omega (\bm a + \bm b z)]
	~ ~
	\text{s.t.}
	~ ~
	 S \bm t = \bm h, \bm t \geq \bm 0 \\ 
	 \Leftrightarrow
	 & \min \limits_{\bm t \in \RR^{nm}} ~  
	(\bm u + \bm v z + \bm w z^2)^\top \bm t
	~ ~
	\text{s.t.}
	~ ~
	 S \bm t = \bm h, \bm t \geq \bm 0,
\end{align*}
where 
\begin{align*}
	\bm u &= (\Omega \bm a) \circ (\Omega \bm a),\\ 
	\bm v & = (\Omega \bm a) \circ (\Omega \bm b) + (\Omega \bm b) \circ (\Omega \bm a), \\ 
	\bm w &= (\Omega \bm b) \circ (\Omega \bm b),
\end{align*}
and $S$ and $\bm h$ are the same as in \eq{eq:wasserstein_reformulated}.
By fixing $\cM_{\rm {obs}}$ as the optimal basic index set, the \emph{relative cost vector} w.r.t to the set of non-basic variables is defines as 
\begin{align*}
	\bm r_{\cM^c_{\rm {obs}}} = \tilde{\bm u} + \tilde{\bm v} z + \tilde{\bm w} z^2,
\end{align*}
where 
\begin{align*}
\tilde{\bm u} &= 
\left(
	\bm u_{\cM^c_{\rm {obs}}}^\top - \bm u_{\cM_{\rm {obs}}}^\top S_{:, \cM_{\rm {obs}}}^{-1} S_{:, \cM^c_{\rm {obs}}}
\right)^\top,
\\ 
\tilde{\bm v} &= 
\left (
	\bm v_{\cM^c_{\rm {obs}}}^\top - \bm v_{\cM_{\rm {obs}}}^\top S_{:, \cM_{\rm {obs}}}^{-1} S_{:, \cM^c_{\rm {obs}}}
\right )^\top,
\\ 
\text{ and } ~~ 	
\tilde{\bm w} &= 
\left (
	\bm w_{\cM^c_{\rm {obs}}}^\top - \bm w_{\cM_{\rm {obs}}}^\top S_{:, \cM_{\rm {obs}}}^{-1} S_{:, \cM^c_{\rm {obs}}}
\right )^\top.
\end{align*}
The requirement for $\cM_{\rm {obs}}$ to be the optimal basis index set is $\bm r_{\cM^c_{\rm {obs}}} \geq \bm 0$ (i.e., the cost in minimization problem will never decrease when the non-basic variables become positive and enter the basis).
Finally, the set $\cZ_1$ can be defined as 
\begin{align*}
	\cZ_1 &= \{ z \in \RR \mid  \cM(\bm a + \bm b z) = \cM_{\rm {obs}}\}, \\ 
	\cZ_1 &= \{ z \in \RR \mid  \bm r_{\cM^c_{\rm {obs}}} = \tilde{\bm u} + \tilde{\bm v} z + \tilde{\bm w} z^2 \geq \bm 0\},
\end{align*}
which can be obtained by solving the system of quadratic inequalities.

\section{Experiment on High-dimensional Data}
\begin{figure}[!t]
\centering
\includegraphics[width=.55\linewidth]{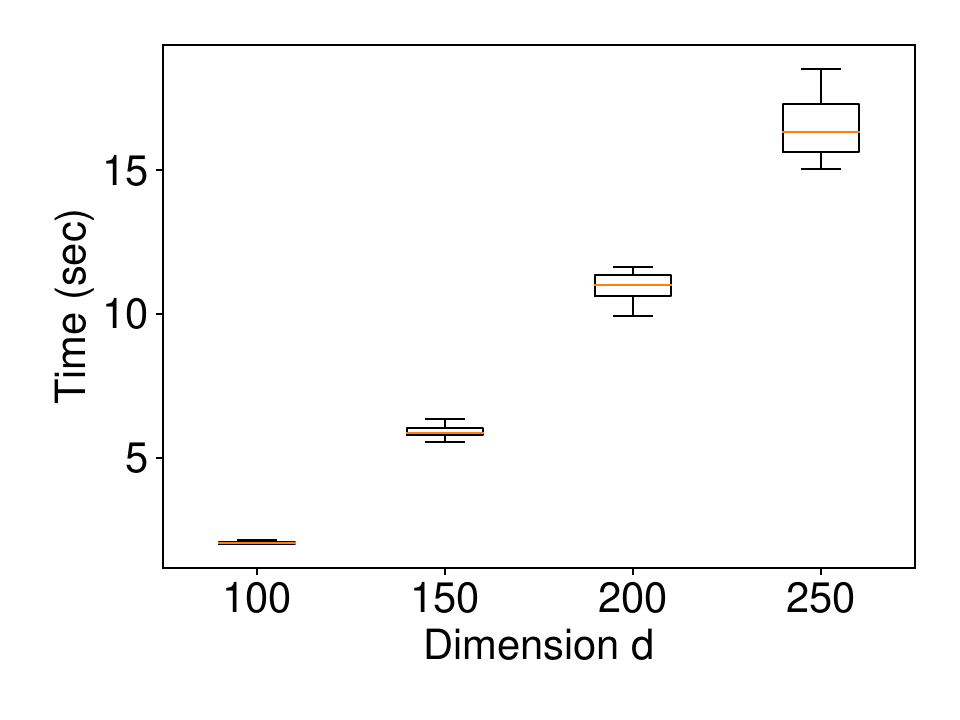}  
\caption{Computational time of the proposed method when increasing the dimension $d$.}
\label{fig:cc_increase_d}
\end{figure}

We generated the dataset 
$X = \{\bm x_{i, :} \}_{i \in [n]}$ 
with $\bm x_{i, :} \sim \NN (\bm 1_d, I_d)$
and 
$Y = \{\bm y_{j, :} \}_{j \in [m]}$ 
with $\bm y_{j, :} \sim \NN (\bm 1_d + \Delta, I_d)$  (element-wise addition).
We set $n = m = 20, \Delta = 2$, and ran 10 trials for each $d \in \{100, 150, 200, 250\}$. 
The results in Figure \ref{fig:cc_increase_d} show that
the proposed method still has reasonable computational time. 

\clearpage

\end{document}